# Fitness aligned structural modeling enables scalable virtual screening with AuroBind

Zhongyue Zhang[#,1,2], Jiahua Rao[#,1,3], Jie Zhong[#,4], Weiqiang Bai[#,5,6], Dongxue Wang[#,7], Shaobo Ning[4], Lifeng Qiao[6], Sheng Xu[5,6], Runze Ma[1,2], Will Hua[2], Jack Xiaoyu Chen[8], Odin Zhang[1], Wei Lu[1], Hanyi Feng[7], He Yang[4], Xinchao Shi[4], Rui Li[4], Wanli Ouyang[6], Xinzhu Ma[6], Jiahao Wang[1,2], Jixian Zhang[1], Jia Duan[7*], Siqi Sun[5,6*], Jian Zhang[4,9*], Shuangjia Zheng[1*]

[1]Global Institute of Future Technology, Shanghai Jiao Tong University, Shanghai, China
[2]Lingang Laboratory, Shanghai, China
[3]School of Computer Science and Engineering, Sun Yat-sen University, Guangdong, China
[4]Medicinal Chemistry and Bioinformatics Center, School of Medicine, Shanghai Jiao Tong University, Shanghai, China
[5]Research Institute of Intelligent Complex Systems, Fudan University, Shanghai, China
[6]Shanghai Artificial Intelligence Laboratory, Shanghai, China
[7]Zhongshan Institute for Drug Discovery, Shanghai Institute of Materia Medica, Chinese Academy of Sciences, Guangdong, China
[8]Institute for Medical Engineering & Science, Department of Biological Engineering, Massachusetts Institute of Technology, Cambridge, MA, USA
[9]Key Laboratory of Protection, Development and Utilization of Medicinal Resources in Liupanshan Area, Ministry of Education, Peptide & Protein Drug Research Center, School of Pharmacy, Ningxia Medical University, Ningxia, China

[#]These authors contributed equally.
[*]Correspondence to: shuangjia.zheng@sjtu.edu.cn; jian.zhang@sjtu.edu.cn; siqisun@fudan.edu.cn; duanjia@simm.ac.cn

## Abstract

Most human proteins remain undrugged, over 96% of human proteins remain unexploited by approved therapeutics. While structure-based virtual screening promises to expand the druggable proteome, existing methods lack atomic-level precision and fail to predict binding fitness, limiting translational impact. We present AuroBind, a scalable virtual screening framework that fine-tunes a custom atomic-level structural model on million-scale chemogenomic data. AuroBind integrates direct preference optimization, self-distillation from high-confidence complexes, and a teacher–student acceleration strategy to jointly predict ligand-bound structures and binding fitness. The proposed models outperform state-of-the-art models on structural and functional benchmarks while enabling 100,000-fold faster screening across ultra-large compound libraries. In a prospective screen across ten disease-relevant targets, AuroBind achieved experimental hit rates of 7–69%, with top compounds reaching sub-nanomolar to picomolar potency. For the orphan GPCRs GPR151 and GPR160, AuroBind identified both agonists and antagonists with success rates of 16–30%, and functional assays confirmed GPR160 modulation in liver and prostate cancer models. AuroBind offers a generalizable framework for structure–function learning and high-throughput molecular screening, bridging the gap between structure prediction and therapeutic discovery.

**Main Text**

The human genome encodes approximately 20,000 protein-coding genes, many of which are implicated in disease. However, only about 3–4% of these genes currently have drug candidates available for therapeutic intervention[1]. While experimental high-throughput screening methods, including combinatorial libraries[2], DNA-encoded libraries[3-5] and mass spectrometry-based screening[6,7], have enabled hit discovery in some cases, they remain labor-intensive, costly, and difficult to direct toward specific binding pockets.

Structure-guided virtual screening approaches offer a scalable alternative[8]. Physics-based methods like Glide[9] and AutoDock Vina[10] have played a key role in early hit discovery, but their performance declines for convex or highly polar binding sites, often yielding weak binders and failing on intractable targets. Moreover, their reliance on high-resolution bound (holo) structures and predefined binding sites limits applicability to cryptic pockets[11]. Deep learning–based predictors for protein–ligand complex structures and fitness have recently emerged[12-18], but the accuracy of such attempts remains mixed and often falls short of physics-based methods[19]. Importantly, their real-world utility in prospective virtual screening remains largely untested. These limitations underscore the need for more accurate, scalable computational frameworks that bridge structural modeling and therapeutic validation.

Generative foundation models for biomolecular interactions, such as RoseTTAFold All-Atom[20] and AlphaFold 3[21], offer new opportunities for structure-based drug discovery. These models encode deep structural and evolutionary priors but face two major limitations when applied to ligand discovery. First, these models lack fitness-aware training, making it difficult to distinguish true hits from decoys. Second, they require substantial inference time per sample, limiting their scalability to screen ultra-large chemical libraries. We reason that extending generative foundation models with functional awareness and scalable screening capabilities, while preserving their structural fidelity, would revolutionize molecular discovery landscape and enable more systematic exploration of druggable targets across the proteome.

To bridge these gaps, we introduce AuroBind, a generative framework that bridges structure prediction and binding fitness modeling for efficient virtual screening. AuroBind integrates three key elements: (i) preservation of AlphaFold 3-level structural accuracy through training on a curated semi-synthetic dataset of 330k protein-ligand complexes, (ii) functional fitness alignment via fine-tuning on ~1.27 million chemogenomic measurements across 1,300 diverse protein domain; and (iii) a two-stage distillation pipeline that produces a lightweight, ultra-fast screening module capable of navigating chemical libraries at scale. AuroBind achieves state-of-the-art performance across a wide range of virtual screening and experimental benchmarks, outperforming both physics-based and deep learning models in structure prediction, enrichment, and computational efficiency.

To validate AuroBind experimentally, we performed a systematical virtual screening campaign to screen a 30-million-chemical library against 10 diverse, challenging and therapeutically relevant protein targets, including receptor tyrosine kinases, epigenetic regulators and understudied G protein-coupled receptors (GPCRs). The whole virtual screening process can be finished within 24 hours on a local HPC cluster equipped with two H800 GPUs for each target. Across these targets, AuroBind achieved experimental hit rates ranging from 7–69%, with top hit compounds reaching sub-nanomolar to picomolar potency. More importantly, AuroBind demonstrated strong generalization to orphan GPCRs, identifying potent agonists and antagonists without requiring any known binders, co-crystal structures, or predefined binding pockets. These results establish AuroBind as a scalable and effective platform for the on-demand discovery of high-fitness small molecules that functionally modulate diverse protein targets.

**Results**

**AuroBind enables joint structure–fitness learning and scalable screening**

AuroBind takes as input a protein sequence and a ligand SMILES (simplified molecular-input line-entry system), and jointly predicts the 3D structure of the ligand-bound protein complex along with a scalar fitness score reflecting binding potency. The model is designed to support atomic-resolution, high-throughput virtual screening across ultra-large chemical libraries. AuroBind builds on an architecture inspired by AlphaFold 3 but optimized for binding fitness prediction, leveraging two critical capabilities of structural foundation models: the ability to model precise protein-ligand physical interactions and the capacity to incorporate co-evolutionary signals from multiple sequence alignments.

The framework retains the general layout of AlphaFold 3 **(Fig. 1a; Methods)**, with three key extensions: (1) a chemogenomic embedding module and fitness prediction head for predicting both per-residue and global fitness scores; (2) a self-distillation module that refines structural accuracy using high-confidence, high-fitness complexes; and (3) a lightweight student model, AuroFast, that accelerates inference by orders of magnitude without sacrificing accuracy. Together, these elements allow AuroBind to predict both atomic-level binding poses and fitness in a single forward pass, while efficiently scaling to millions-scale compound libraries (**Fig. 1b**).

Training of AuroBind proceeds in two stages. In the first stage, we focused on protein–ligand structure prediction. We curated a dataset of approximately 100,000 protein–ligand complexes derived from the Protein Data Bank (PDB)[22] for this initial supervised training. To further improve generalization and structural fidelity, we employed a self-distillation strategy: the initially trained model generated predicted structures protein–ligand structures on the BindingDB dataset, which contains nearly 500,000 complexes, after excluding pairs with high fitness and confidence, this filtering yielded 230,000 self-distilled protein–ligand complexes for subsequent training **(Supplementary Fig. 1; Methods)**.

The second stage involves fine-tuning the structure model from stage one on a large-scale dataset of approximately 1.27 million protein–ligand chemogenomic pairs curated from ChEMBL[23] and PubChem[24] following the data cleaning strategies by ExcapeDB[25], **(Methods)**. This dataset encompasses a diverse range of protein targets, small molecules and their corresponding scalar-labeled binding fitness (pXC50 value). A fitness prediction head was appended after Pairformer to estimate binding fitness from the learned protein–ligand representations **(Fig. 1c, SI Methods 1.2.1)**. Initially, the model was trained using a mean squared error (MSE) objective to establish baseline fitness prediction capability, and once stabilized, we leveraged a direct preference optimization (DPO)[26] objective to further fine-tune the model **(Methods)**. During DPO, structural confidence scores served as soft weights, encouraging the model to prioritize high-confidence structure–function associations while mitigating noise from weak or ambiguous binding data **(Methods)**. This strategy sharpens the model's ability to distinguish between ligands with subtle differences in binding fitness—an essential capability for effective prioritization in large-scale virtual screening.

To support ultra-high-throughput virtual screening, we distill a lightweight student model, named AuroFast, from AuroBind via a teacher–student paradigm **(Methods)**. AuroFast maintains the ability to predict both structural embedding and fitness, but operates at ~100,000× faster inference speed, enabling the screening of 20–30 million compounds per target within hours on standard GPU clusters.

During inference, AuroBind follows a hierarchical screening protocol: AuroFast is first used to screen the full chemical library and prioritize thousands of candidate compounds with high predicted fitness. This subset is then re-evaluated by AuroBind to generate full ligand–protein complex structures and refined fitness scores. After that we performed post-processing including drug-likeness filtering, structure evaluation and commercial availability screening, yield a final compound set for prospective wet-lab validation **(Fig. 1d; Methods)**. This two-stage approach draws inspiration from classical docking pipelines, which use a coarse-grained scoring function for initial triage, followed by higher-resolution rescoring to refine candidate selection[27,28].

**Accurate and scalable prediction across fitness and structure benchmarks**

An ideal method for virtual screening should enable i) accurately predicting binding fitness, ii) scaling efficiently to ultra-large chemical libraries, and iii) capturing atomic-level protein–ligand structures. To this end, we sought to evaluate AuroBind across fitness ranking, screening efficiency, and structural accuracy, using a wide range of computational benchmarks.

We first benchmarked AuroBind's ability to prioritize compounds by binding fitness, a critical step for experimental follow-up. On the DAVIS[29] and BindingDB[30] datasets, we compared against AlphaFold 3[10], Ridge regression, ConPLex[17], MolTrans[31], DeepConv-DTI[32], EnzPred CPI[33] and GNN-CPI[34]. All models were evaluated under the same training split as defined in the ConPLex[17] framework, which explicitly excludes any complexes shared between the training and test sets to prevent data leakage. While AlphaFold 3 does not directly model fitness, its structure confidence score can serve as a weak proxy, achieving area under the precision-recall curve (AUPR) scores of 0.08 (DAVIS) and 0.13 (BindingDB). AuroBind-ZS, a zero-shot variant not fine-tuned on these datasets, improved performance to 0.22 and 0.24, respectively, the best model that did not train directly on these datasets. When fine-tuned (AuroBind-FT), the model achieved AUPR of 0.61 on DAVIS and 0.70 on BindingDB, representing significant improvements of 33.8% and 11.6% over the best supervised baseline (ConPLex) (**Fig. 2a, b**). In addition to predictive accuracy, AuroBind provides residue-level interpretability through spatial fitness mapping. In the protein–ligand complex with PDB ID 7A1P, high-scoring regions align with hydrogen-bonding hotspots predicted by plip, suggesting the model can capture physicochemical determinants of molecular recognition (**Fig. 2c**).

We further evaluated the performance of model on large-scale virtual screening. The distilled version AuroFast was evaluated in a zero-shot setting on the largest virtual screening benchmark LIT-PCBA[35], comprising over 2.6 million protein–ligand complexes across 15 targets **(see SI Methods 1.5)**. To avoid data leakage, we exclude all the complex present in LIT-PCBA from our training set for both AuroBind and AuroFast. This approach aligns with established zero-shot evaluation protocols, as demonstrated in recent studies[36]. Compared to deep learning-based ultra-fast screening methods (DrugCLIP[37], GNINA[38], Planet[39], BigBind[36], and DeepDTA[40]) and classical docking approaches (Glide-SP[9] and Surflex[41]), AuroFast achieved an enrichment factor 1% score of 7.58, exceeding the top-performing baseline by 37.6% (**Fig. 2d**). Notably, AuroFast enables screening throughput of >100,000× faster than AlphaFold 3 and >25,000× faster than AutoDock Vina, making it feasible to pre-filter tens of millions of compounds per target within hours on standard GPU hardware (**Fig. 2e**). We did not include generative structure predictors such as AlphaFold 3 due to their low computational efficiency when scaled to million-level compound libraries.

We next assessed whether AuroBind maintains high-resolution structural prediction of protein–ligand complexes after being fine-tuned for functional accuracy. To ensure a fair evaluation, PoseBusters[42] analysis was performed with a training cut-off of September 2021 for AuroBind, excluding all PoseBusters structures from their training sets (**see SI Methods 1.5**). On the PoseBuster V1 and V2 benchmarks, AuroBind achieved success rates of 79.1% and 81.7%, outperforming AlphaFold 3 (78.0% and 81.0%) and Protenix[43] by larger margins (6.0% and 6.8%, with McNemar p-value<0.004) (**Fig. 2g, h**). These results align with our expectations, as the AuroBind was self-distilled with supplementary high-quality structural data. Compared to classical docking tools like AutoDock Vina and GOLD, AuroBind improved structural accuracy by 26.8–27.9% (PoseBuster V1) and 22.1–23.6% (PoseBuster

V2), particularly in flexible binding pockets. These global metrics were mirrored at the local level. As an example in the 7OFF complex, AuroBind accurately recovered key polar contacts observed in the crystal structure, including a hydrogen bond network among GLU106, ASP141, and ARG139. In contrast, AlphaFold 3's predicted pose missed several critical interactions, reducing structural plausibility (**Fig. 2f**).

Together, these results demonstrate that AuroBind achieves state-of-the-art performance across all virtual screening and structure prediction benchmarks, while no previously reported model consistently performs at a top level across these tasks.

**Systematic virtual screening across diverse protein targets with unprecedented success rates**

To validate AuroBind experimentally, we conducted a systematic wet-lab screening campaign across ten protein targets spanning a range of structural classes and therapeutic areas. The selected targets include receptor tyrosine kinases (TrkB, HER3), serine kinases (GSK3α, GSK3β, CDK2), epigenetic regulators (HDAC3), and G protein-coupled receptors (CCR4, mGluR5, GPR151, GPR160). Target selection was guided by a combination of biological relevance, assay feasibility, and the desire to span a gradient of virtual screening challenges (SI Methods **2.1**). Specifically, we included well-characterized drug targets with established assays (e.g., kinases and HDACs) to benchmark AuroBind against prior methods, alongside more challenging GPCRs (CCR4 and mGluR5). After experimental validation on the first 8 targets was completed, two orphan GPCRs (GPR151 and GPR160) were selected to test the model's ability to generalize to low-data, structurally intractable targets. While we did not anticipate that all targets would yield high-confidence hits, the observed success across diverse target classes demonstrate the generality and robustness of AuroBind.

For each target, we applied a hierarchical screening protocol (**Fig.3a**). In stage 1, AuroFast was used to perform rapid fitness-based pre-filtering across libraries exceeding 30 million purchasable compounds, sourced from the ZINC, commercial off-the-shelf drug library[44] and MedChemExpress (MCE) libraries[45]. In stage 2, the 10,000 top-ranking molecules from the AuroFast were re-scored by AuroBind to obtain high-confidence ligand–protein complex structures and refined fitness scores. The top-ranked 500 compounds were then automatically filtered based on drug-likeness and predicted solubility to yield a final selection. To ensure structural novelty, we removed compounds with high similarity (a cut off of Tanimoto similarity>0.6) to known actives for each target recorded in ChEMBL[23]. Notably, we performed minimal binding mode–based filtering, as our screening protocol does not rely on predefined binding pockets though it supports binding site-based screening **(see Methods)**.

We prioritized approximately 50 compounds per target for experimental testing. Primary screening was performed at a concentration of 10 μM in biological triplicates (n = 3 per compound) based on different biochemical and functional assays. Experimental validation was performed using biochemical or biophysical assays appropriate to each target class, including kinase activity inhibition, TR-FRET displacement, glosensor (cAMP inhibition) and BERT2 assays (**SI Methods 2.4**).

Due to differences in compound availability and solubility, an average of 30–50 compounds per target were ultimately validated. Across the ten targets, we observed experimental hit rates ranging from 7% (HDAC3) to 69% (GSK3α) at a concentration of 10 μM (**Fig.3b**). Per-target hit rates were >10% for 8 targets, >20% for 5 targets and >30% for 4 targets. Notably, for GPR151 and GPR160, AuroBind is able to identify both agonists and antagonists, despite the absence of previously reported active compounds or crystal structures and low homology (sequence identity<0.3) with any protein in the fitness training set. On GSK3α, HER3, CDK2 and mGluR5AuroBind achieved 49-fold, 69-fold, 324-fold and 1474-fold higher hit rates, respectively, compared to the previously reported results (**Supplementary Table 4**). These results show that AuroBind can reliably support large-scale virtual screening and identify active compounds across a wide range of protein targets, including those that are poorly characterized.

**Scalable discovery of potent, diverse and unseen functional hits**

While high experimental hit rates reduce the cost and effort of identifying active compounds, downstream utility in drug discovery depends critically on three additional metrics: binding fitness, chemical diversity, and molecular novelty. To evaluate binding fitness, we selected the top 3–5 most active compounds per target for dose–response assays. AuroBind prioritized functionally active compounds with nanomolar to sub-nanomolar potency: $IC_{50}$ values below 1 nM were observed for 2 targets, below 10 nM for 5 targets and below 1 μM for 8 targets (**Fig.3b, 4a**, **Supplementary Fig. 3, 4**).

The strongest observed hit was a TrkB-targeting compound with a $IC_{50}$ of 220 pM (**Fig.3b**). In total, three sub-nanomolar hits were identified, including two for TrkB and one for GSK3α. Compared to the best unoptimized hit compounds from prior screening methods, AuroBind yielded superior $IC_{50}$ values across all tested targets—outperforming by factors of 15.3× (mGluR5), 435× (HER3), 818× (TrkB) and 4185× (GSK3α) (**Supplementary Table 4, Fig. 3**). Even relative to previous hit compounds that were iteratively optimized in wet-lab cycles, AuroBind still produced better or comparable potency on CCR4 (6.78×) and HDAC3 (286×) (**Supplementary Table 4**, "Optimized in wet-lab cycles"). These results suggest that AuroBind can discover high-potency hit compounds for diverse targets after a single virtual screening round of 30–50 candidates, without the need for subsequent optimization.

To better assess the correlation between predicted fitness and experimental fitness, we further evaluated the enrichment performance of AuroBind models using the enrichment factor at 1% (EF1%) under a 10 μM inhibition threshold. For each target, we designated as negatives all ChEMBL compounds with reported activity > 20 μM and in-house experimentally tested molecules that demonstrated less than 50% inhibition, resulting in target-specific negative-to-positive ratios that varied across the dataset (**see SI Methods 1.5, Supplementary Table 3**). We compared AuroBind, AuroFast, confidence scores from AlphaFold 3, as well as a physics-based docking baseline Vina. Hit identification was performed at two inhibition thresholds: >50% and >90% at 10 μM. Across all benchmarks, AuroBind significantly outperformed other approaches in EF1%, indicating superior ability to enrich true positives from large chemical libraries (**Fig.4b**). These results indicate that AuroBind is strongly competitive to deep-learning and physics-based screening methods in terms of hit enrichment.

We also evaluated the chemical diversity and novelty of the identified hit compounds. To assess scaffold-level diversity, we randomly selected 10,000 cluster centers from a 30 million–compound chemical library and projected AuroBind-identified hits from each target onto this chemical space. In all cases, the hit compounds exhibited broad and evenly distributed coverage across diverse chemical scaffolds (**Fig.4 c, 4d**). To validate the chemical novelty of the screened compounds, we computed structural similarity to known active molecules from CHEMBL and observed a gradient distribution (**Fig.4 e**), with 25% (71/283) compounds exhibiting Tanimoto similarity < 0.30 to any previously reported active compound. The gradient demonstrates the dual strengths of our model by showing its ability to uncover structurally diverse compounds for prospective discovery, and its capacity to retrieve molecules similar to known actives, confirming retrospective performance. This supports the conclusion that AuroBind discovers structurally diverse molecules beyond the reach of traditional methods. Remarkably, AuroBind also identified multiple potent hits for GPR160 and GPR151, which no active small molecules or structural templates had been previously reported. This demonstrates the model's ability to generalize to targets that are both undrugged and structurally uncharacterized, a setting where physics-based docking methods often fail due to the absence of reliable structural inputs.

**Targeting undruggable orphan GPCR with AuroBind**

Structural predictions from AlphaFoldDB[46] suggest that GPR151 adopts a relatively canonical GPCR conformation, whereas GPR160 deviates substantially, exhibiting atypical architecture and marked conformational flexibility, posing significant challenges for ligand docking (**Supplementary Fig. 5**). To evaluate the generalizability of AuroBind on previously intractable targets, we focused on GPR151[47] and GPR160[48], two orphan receptors lacking known ligands and experimentally determined structures, yet implicated in diseases such as neuropathic pain, diabetes, and cancer[49-52].

From the top-ranked predictions, we tested ~50 commercially available compounds for functional validation using BRET2[53] and GloSensor assays[54], respectively (**see SI Methods 2.4.5, 2.4.6**) as GPR151 and GPR160 primarily signal through the Gi/o pathway (GPR151 via Go, GPR160 via Gi)[55]. Remarkably, in the initial screening, 7 of 42 compounds targeting GPR151 showed either agonistic or antagonistic activity (**Fig. 5a**). Among them, GPR151-C15 and GPR151-C40 elicited robust Go protein recruitment, with $EC_{50}$ values of ~1100 nM and ~4200 nM, respectively (**Figs. 5b, c**). Structural modeling confirmed that both ligands bind within the orthosteric pocket (**Fig. 5d**). Detailed analysis revealed that both compounds form extensive hydrophobic and polar interactions with GPR151, engaging the receptor N-terminus, transmembrane helices TM1–3, TM5–7, and extracellular loops ECL1 and ECL2 (**Figs. 5d, e**). Compared to GPR151-C40, GPR151-C15 forms additional contacts with residues V189$^{ECL2}$, W267$^{6.52}$, and I288$^{7.35}$—likely contributing to its greater potency and efficacy (**Figs. 5e**).

In contrast, GPR160 displays a markedly noncanonical conformation. Screening identified 14 out of 46 compounds with either agonistic or antagonistic activity (**Fig. 5f**). Notably, compounds GPR160-C45 and GPR160-C05 significantly activated GPR160, with EC$_{50}$ values of ~1200 nM and ~1250 nM, respectively (**Figs. 5g, h**). Interestingly, the two agonists appear to bind at distinct sites, underscoring AuroBind's strong potential in identifying diverse ligandable pockets (**Figs. 5i**). These contrasting models underscore the need for future high-resolution structural studies to clarify the ligand-binding mechanisms of GPR160. Together, these findings demonstrate that AuroBind achieves high hit rates and effectively identifies functionally active ligands—even for structurally atypical or previously uncharacterized orphan GPCRs like GPR160. In summary, AuroBind provides a powerful approach for ligand discovery across a broad spectrum of orphan GPCRs, including those traditionally considered undruggable.

**Discussion**

A central challenge in structure-based drug discovery is to identify small molecules that modulate the biological function of underexplored protein targets. Our study demonstrates that functionally fine-tuning generative structural foundation models enables a scalable and accurate virtual screening framework capable of retrieving diverse, potent functional hits across a wide range of target classes. Specifically, AuroBind achieves this by shifting from purely geometric interaction modeling toward function-aware structure learning, capturing both physicochemical plausibility and functional relevance.

AuroBind addresses long-standing limitations in traditional docking and deep learning–based pipelines, which often struggle with low hit rates, restricted generalizability, and limited scalability. In prospective evaluations across ten biologically and structurally diverse targets, AuroBind achieved hit rates ranging from 7% to 69%, substantially higher than typical docking pipelines. Several of the identified compounds demonstrated picomolar potency, and 25% of them showed low structural similarity to known actives, demonstrating the framework's ability to explore sparsely sampled regions of chemical space and prioritize functional candidates. Importantly, AuroBind does not require experimentally resolved complex structures to succeed. For instance, despite the absence of structural templates and ligand data for GPR160, a conformationally flexible orphan GPCR, AuroBind identified multiple potent hits, supporting the model's robustness in structurally ambiguous settings.

In contrast to prior models that rely on ligand similarity or rigid geometric fit, AuroBind is trained to infer functionally meaningful interactions from sequence and ligand structure alone. This inductive bias allows the model to generalize to previously unseen targets, including orphan receptors and poorly characterized enzymes. By integrating structural representation learning with chemogenomic fitness data, AuroBind moves beyond pose prediction to capture structure–function relationships critical for drug discovery.

Nonetheless, several limitations remain. First, the model's performance on highly dynamic or disordered targets, such as KRAS or transcription factors like c-Myc, requires further evaluation. Second, while AuroBind demonstrates robust functional generalization, the accuracy of predicted fitness scores still leaves room for improvement, especially in sparse-data regimes where fine-grained distinctions may be subtle. Third, while the model can generalize in zero-shot settings, additional strategies may be needed to improve performance on extremely sparse protein families.

It is worth noting that during the review process of this manuscript, several contemporaneous efforts—such as Boltz-2[56]—have been posted as preprints, similarly exploring the use of structural foundation models for fitness prediction. This parallel emergence underscores a growing convergence in the field toward function-informed structural learning. What distinguishes our work, however, is threefold: (1) support for ultra-high-throughput virtual screening at scale, (2) the establishment of a closed loop from in silico prediction to prospective experimental validation, and (3) the ability to discover hits for unseen, structure-unknown targets. Through rigorous wet-lab validations across diverse targets, we demonstrate that AuroBind not only generalizes across protein families but also consistently identifies potent functional hits, thereby underscoring its practical utility in real-world drug discovery.

Overall, our results highlight the potential of functionally fine-tuned structural foundation models to bridge the gap between geometry and biology. As generative modeling techniques and structural databases continue to evolve, frameworks like AuroBind may serve as a foundation for autonomous, high-throughput, and function-informed drug discovery, bringing us closer to AI-driven platforms capable of end-to-end therapeutic design.

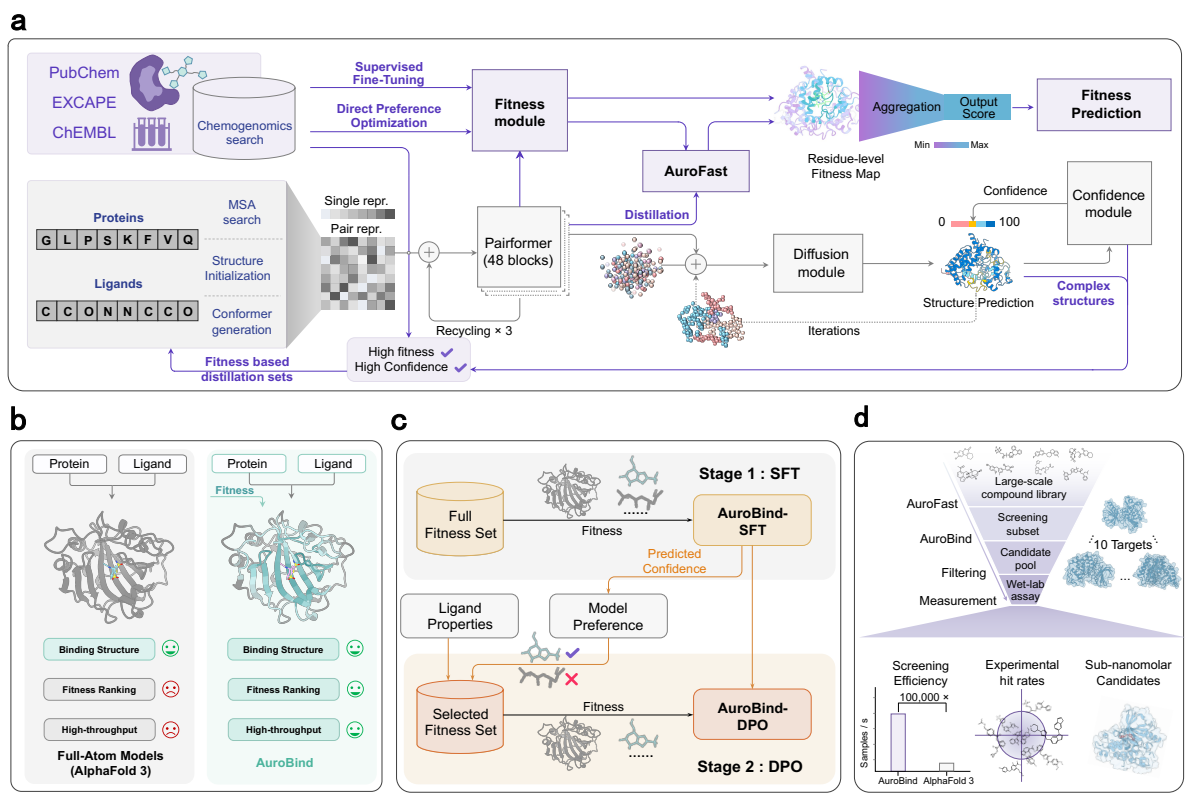

**Fig. 1 | Overview of the AuroBind framework for protein–ligand structure and fitness prediction.**

**a,** Model architecture. AuroBind takes protein sequences and ligand SMILES as input. It performs multiple sequence alignments (MSAs) search for the protein and generates ligand conformers. A 48-block PairFormer captures joint protein–ligand representations, which are passed to a diffusion module to generate atomic-resolution complex structures. A fitness module predicts the residue-level fitness map and a global binding fitness score for each pair. AuroFast, a distilled student model, enables ultra-fast screening by focusing solely on binding fitness prediction.

**b,** Comparison to a representative structure predictor, AlphaFold 3. While AlphaFold 3 can accurately predict protein–ligand complex structures, it lacks mechanisms for fitness estimation and is computationally intensive for large-scale screening. AuroBind extends this capability by enabling structure-aware fitness prediction and scalable inference through AuroFast.

**c,** Two-stage training strategy. The AuroBind fitness module is optimized in two stages: Stage I involves supervised fine-tuning on ~1.27 million chemogenomic datapoints; Stage II applies direct preference optimization (DPO) on confident structure–fitness pairs, guiding the model to capture functional preferences conditioned on accurate binding geometries.

**d,** High-throughput experimental validation. AuroFast screens millions of compounds across 10 targets. Top-ranked candidates are re-evaluated with full AuroBind predictions. Subsequent experimental screening and dose–response assays confirm high hit rates, strong early enrichment, and sub-nanomolar leads across multiple protein classes.

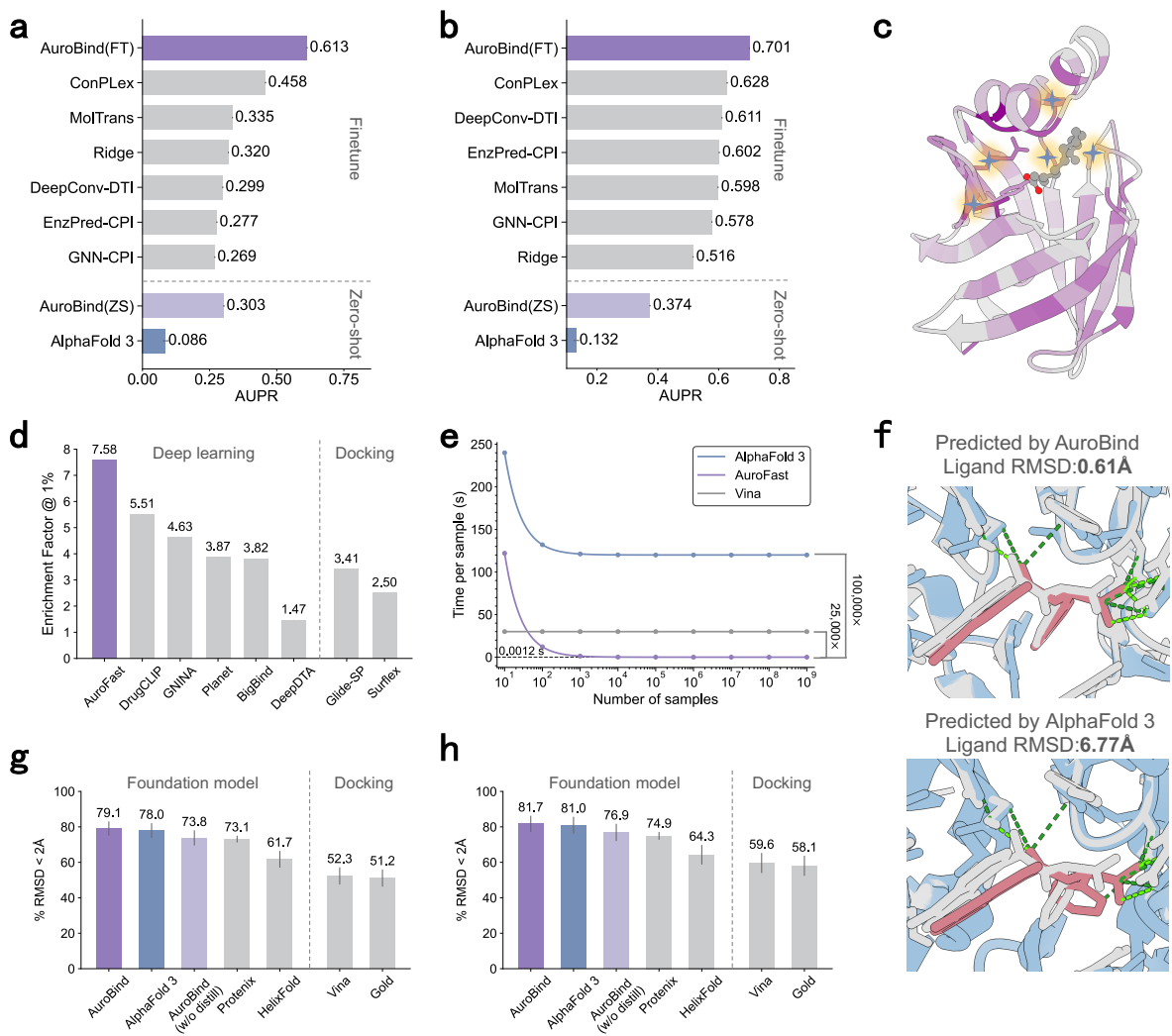

**Fig. 2 | Computational performance of AuroBind across structure, fitness, and screening benchmarks.**

**a,b,** Performance of AuroBind and baseline models on the Davis and BindingDB datasets, measured by area under the precision–recall curve (AUPR). AuroBind outperforms both structure-based and sequence-based methods, including zero-shot and fine-tuned variants.
**c,** Visualization of per-residue binding importance predicted by AuroBind on the 7A1P complex. Darker regions correspond to higher predicted contributions to binding fitness. Star-marked residues indicate known hydrogen bond interaction sites, which align with peaks in predicted importance.
**d,** Early enrichment performance of AuroFast on the LIT-PCBA benchmark, evaluated by enrichment factor at 1% (EF1%). AuroFast outperforms deep learning- and docking-based methods.
**e,** Computational efficiency comparison. Inference time per sample is shown for AuroFast, AlphaFold 3, and AutoDock Vina. AuroFast achieves an average runtime of ~0.0012s per compound at scale, representing a 100,000-fold acceleration over AlphaFold 3 and 25,000-fold over Vina.

**f,** Predicted protein–ligand complex structures for the 7OFF complex, comparing AuroBind (top) and AlphaFold 3 (bottom). Protein chains are shown in blue, predicted ligands in red, and crystallographic ground truth in grey. AuroBind yields lower RMSD and more accurate pose placement.

**g,h,** Success rate comparison on PoseBuster V1(N=428) and V2 benchmarks (N = 308). AuroBind consistently outperforms AlphaFold 3, Protenix, HelixFold, and classical docking tools in predicting binding structures.

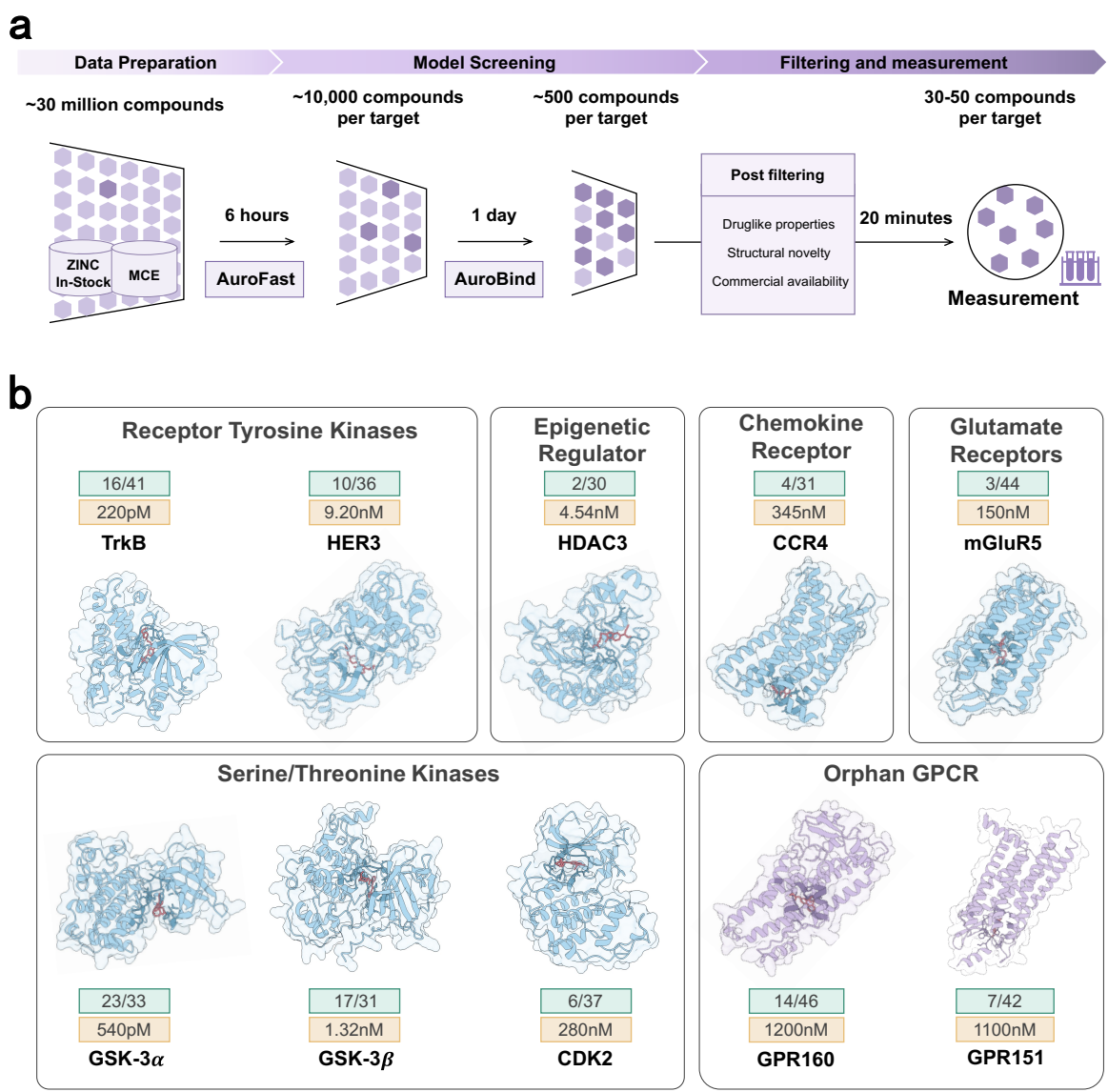

**Fig. 3 | Overview of screening pipeline and experimental performance of AuroBind.**
a, Schematic of the virtual screening and validation workflow. Starting from ~30 million purchasable molecules (sourced from ZINC and MCE), a rapid single-pass fitness screen with AuroFast (~ 6 h per target) reduces each library to ~10,000 candidates. These are re-ranked by full protein–ligand complex predictions with AuroBind (~24 h per target), yielding ~500 top-scoring compounds per target. Post-processing filters, including drug-likeness criteria, structural novelty, and commercial availability, select a final set of 30–50 compounds for wet lab validation.
b, Experimental validation across ten diverse and challenging protein targets. For each target, top-ranked compounds from a multi-million compound library were selected and assayed at 10 μM in biological triplicate. Bars represent the number of experimentally confirmed hits (green) and their corresponding binding affinities (beige). Values in the yellow box indicate the measured $IC_{50}$ of the highest fitness hit without optimization. Hit rates denote the fraction of active compounds (>50% inhibition or activation at 10 μM). Where applicable, functional readouts such as receptor activation or signaling inhibition are also reported.

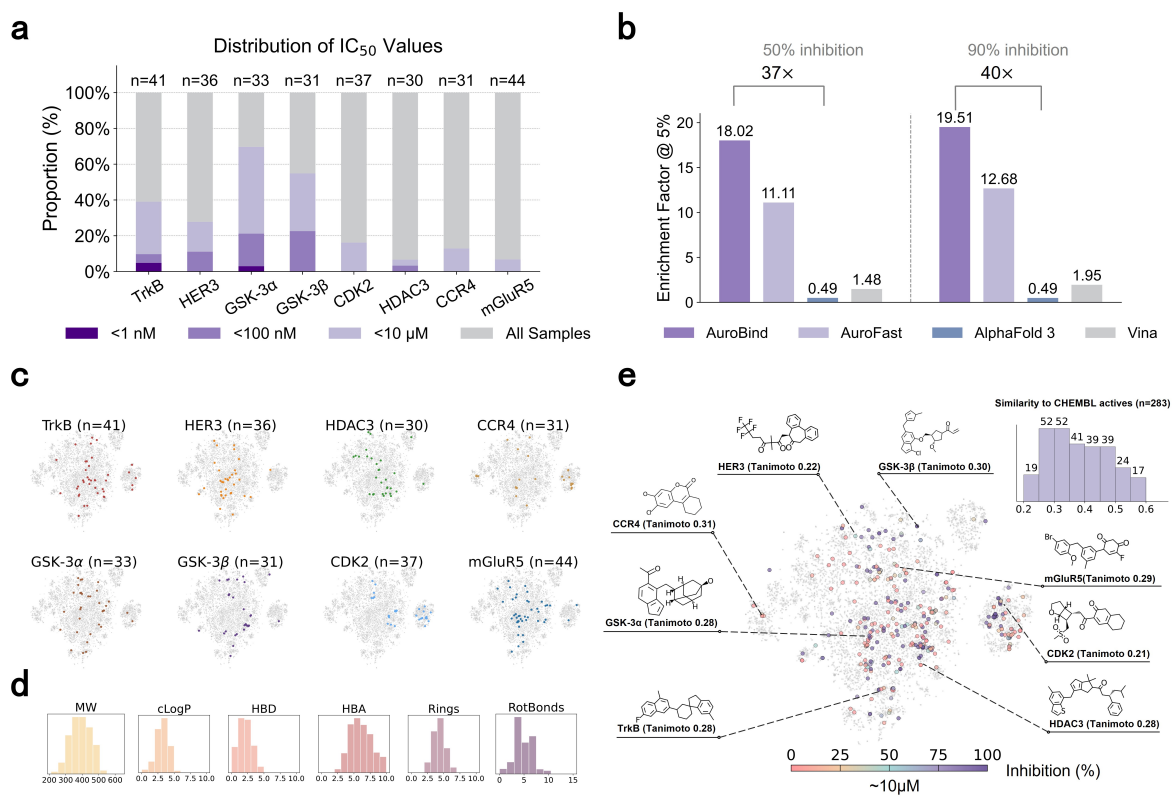

**Fig. 4 | AuroBind identifies potent and structurally novel inhibitors across diverse targets.**

**a,** Proportional distribution of all compounds submitted for experimental testing (n = 30-44 per target) across four potency tiers: <1 nM, <100 nM, <10 μM, and >10 μM (grey, "All Samples"). Bar heights indicate the fraction of compounds within each potency range for the eight protein targets, with stronger hits (darker purple) stacked at the base of each column. Sample counts (n) are shown above the bars.

**b,** Enrichment factor at 5% for compounds with experimental inhibition above 50% (left) or 90% (right), mixing experimentally measured compound and low-fitness decoys from ChEMBL (IC$_{50}$ >20 uM). AuroBind significantly outperforms baseline methods including AuroFast, AlphaFold 3, and Vina.

**c,** t-SNE plots for each target showing the distribution of screened molecules (gray) and experimentally measured compounds (colored), indicating clustering behavior specific to each target.

**d,** Distributions of key physicochemical properties of experimentally measured compounds, including molecular weight (MW), calculated logP (cLogP), number of hydrogen bond donors (HBD), hydrogen bond acceptors (HBA), ring count, and number of rotatable bonds.

**e,** t-SNE visualization of experimentally tested compounds (colored) and the full screening library (gray), with top-ranked AuroBind hits highlighted. For each experimentally tested molecule, its similarity to the closest known ligand in ChEMBL is shown, alongside measured inhibition values. The inset histogram ranks the 283 experimentally confirmed hits by the Tanimoto similarity between each molecule and its closest CHEMBL active (Morgan fingerprints, radius 2).

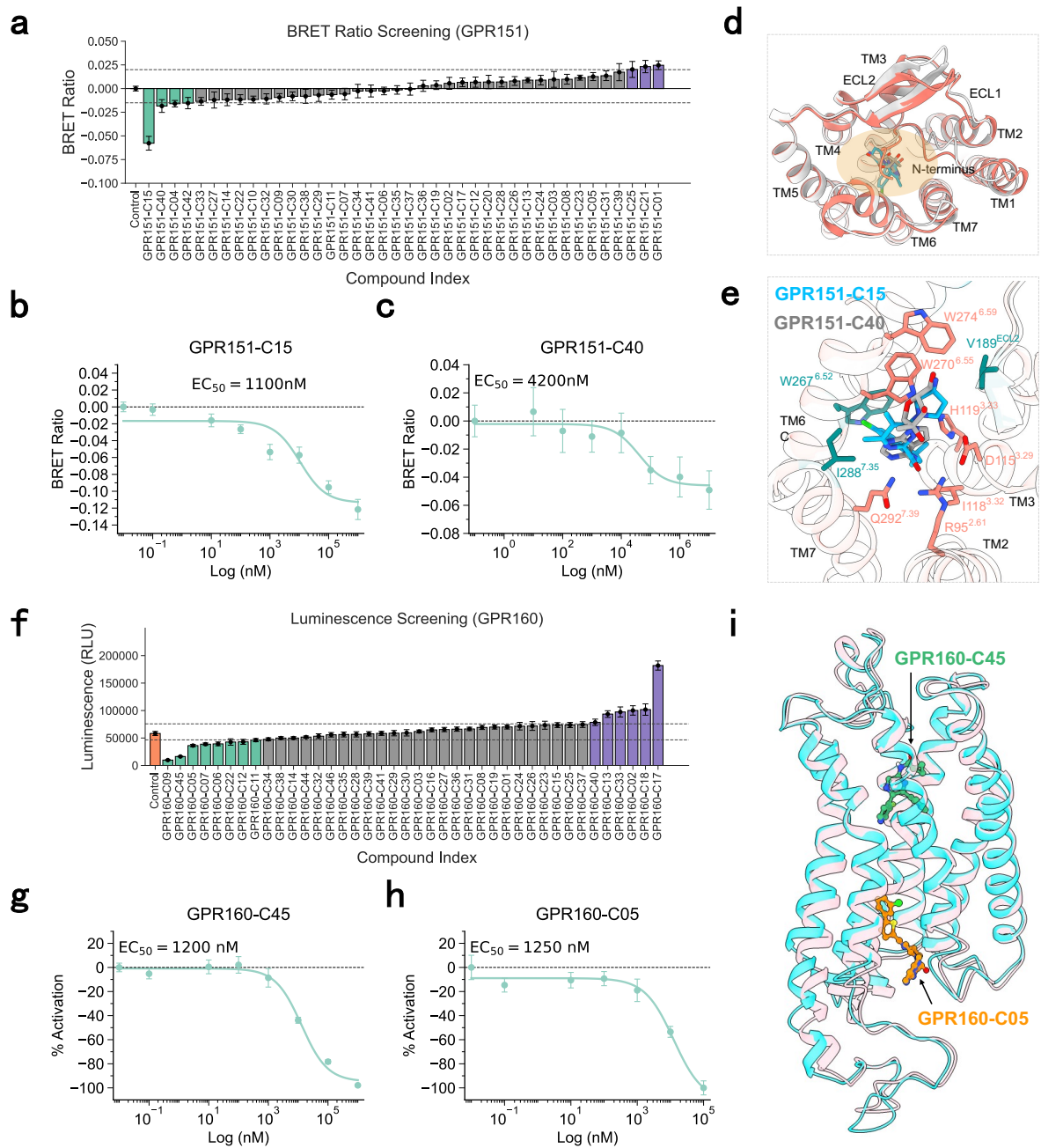

**Fig. 5 | Functional validation of AuroBind-predicted compounds targeting orphan GPCRs.**

**a**, Functional screening results for GPR151 using BRET2 assays. A panel of AI-selected, commercially available compounds was evaluated, yielding several candidates that induced Gi/o-mediated β-arrestin recruitment, with GPR151-C15 showing the strongest signal. +0.02 or -0.015 change in BRET ratio relative to the baseline was used as the corresponding selection threshold, based on levels associated with significant agonistic or inhibitory effects, respectively.

**b-c**, Concentration–response curves for GPR151-C15 (g) and GPR151-C40 (h) on GPR151, plotting % maximal β-arrestin recruitment versus $\log_{10}$[compound] (M). Data are mean ± SEM (n = 3); curves fitted to a four-parameter logistic model yield $EC_{50}$ values of 1100 nM and 4200 nM, respectively.

**d,** Predicted binding poses of GPR151-C15 and GPR151-C40. Both compounds are located within the orthosteric pocket of GPR151, highlighted by an orange circle.

**e,** Detailed interactions of GPR151-C15 and GPR151-C40 with GPR151. Conserved interactions are shown as orange sticks, while compound-specific interactions (for GPR151-C15) are depicted in dark blue.

**f**, GPR160 activation measured via GloSensor cAMP assay revealed two compounds with clear functional activity. Compounds exhibiting a Luminescence decrease greater than 15% (green bars) were classified as putative agonists, whereas compounds with a Luminescence ratio increase greater than 30% (purple bars) were classified as putative inverse agonists or antagonists.

**g-h**, Dose–response validation for the agonists GPR160-C45 and GPR160-C05 confirmed micromolar-range $EC_{50}$ values (~1200 nM and ~1250 nM, respectively). Data points represent mean ± s.e.m. (n = 3). The dashed horizontal lines indicate the baseline (0% activation).

**i**, Structural modeling of GPR160-C45 and GPR160-C05 within the GPR160 ligand-binding pocket reveals distinct binding modes for the two ligands.

## Methods

### Training-data Collection and Curation

Structure prediction training began with ~100,000 high-quality protein–ligand complexes extracted from the PDB (release ≤ 30 September 2021, resolution ≤ 9 Å, ≤ 300 polymer chains). Complexes containing only ions, poorly resolved chains (< 4 residues or all 'UNK'), unstable ligand fragments, or severe atomic clashes (> 30 % of atoms within 1.7 Å of another chain) were discarded, and assemblies with > 20 chains were trimmed to the 20 chains closest to a randomly chosen interface token. After this initial supervision, the partially trained model was used to generate **self-distilled** data: it predicted structures for ~500 000 BindingDB complexes with $IC_{50} \geq 100$ nM, and predictions satisfying ipTM > 0.8, ligand_pTM > 0.5 and pLDDT > 70, while sharing ≤ 60 % sequence identity with any PoseBuster target, were retained, yielding ~230,000 additional complexes. Subsequent epochs were trained on balanced batches of 50,000 examples drawn equally from the original PDB and self-distilled sets. Concentration-response assays from ChEMBL[23] (92,147 assays) and PubChem[24] (58,235 assays) following ExCAPE-DB[25] cleaning rules: active compounds with dose–response ≤ 10 μM and both active and inactive screening records were kept, entries with low-confidence pXC50 = 3.101 or rarely observed molecules were removed, and proteins > 1,000 residues were excluded. Distribution-aware down-sampling yielded a final set of 1,273,781 protein–ligand pairs (1,376 targets, 492,447 unique compounds) with log-transformed activity values. A final direct-preference-optimization pass, applied to pairs whose predicted complex confidence exceeded 0.5, sharpened the model's ability to rank subtle fitness differences.

### Feature Construction

All input complexes were pre-processed with a streamlined workflow adapted from the AlphaFold 3(AF3) pipeline, with every DNA/RNA-specific step removed because AuroBind screens only protein–ligand pairs. For each mmCIF entry we parsed atom coordinates and basic metadata (resolution, release date, experimental method, biological assembly, chemical components and chain information). Multiple-sequence alignments (MSAs) for every protein chain were generated with ColabFold (MMseqs2 + HHblits). Tokens were then created at two granularities: one token per amino-acid residue (Cα as the center atom) and one token per ligand atom; modified residues are tokenized atom-wise.

Three complementary cropping strategies were applied before training. Contiguous cropping selects consecutive residues/atoms following AF3. Spatial cropping retains all tokens within a fixed radius of a randomly chosen reference atom. Spatial-interface cropping chooses the reference atom from an inter-chain interface (≤ 15 Å to another chain's center atom) and again keeps nearby tokens; ligand atoms are always preserved. After cropping, per-token features (indices, chain IDs, type masks), reference-conformer features (3-D coordinates, element, charge, atom-name encodings), MSA-derived features (one-hot sequence, deletion matrices, profiles) and an explicit token–bond matrix are assembled (full list in Supplementary Table 1).

To minimize redundancy, protein chains were clustered at 40 % sequence identity and ligands by exact Chemical Component Dictionary (CCD) identity; no additional interface-level clustering was required because the dataset contains only protein–small-molecule complexes. The resulting, de-duplicated set constitutes the unified training corpus used throughout the structure-prediction and fitness-optimization phases described above.

**AuroBind Architecture**

AuroBind is a structural prediction framework derived from AlphaFold 3 architecture and extended by direct preference optimization on large-scale chemogenomic databases. It enables both fitness prediction and atomic-level protein–ligand structure modeling for accurate, high-throughput virtual screening (Fig. 1a). While retaining the core architecture of AlphaFold 3—with its PairFormer and diffusion modules[10]—AuroBind incorporates an additional fitness prediction module trained using direct preference optimization (DPO) on experimentally validated binding data[48]. This two-stage training strategy allows the model to generate high-quality protein–ligand structures and simultaneously predict the fitness of these interactions, which is crucial for prioritizing candidates in virtual screening.

Like AlphaFold 3, AuroBind accepts protein sequences and ligand SMILES as inputs. For proteins, multiple sequence alignments (MSAs) are first retrieved and processed; for ligands, 3D conformations are initialized with RDKit v.2023_03_3 using ETKDGv3. These inputs are then tokenized and embedded into unified token-level representations. A standard protein residue is represented by a single token, while modified residues and ligands are tokenized at the per-atom level. The PairFormer learns intra- and inter-molecular interactions from these representations, capturing details at both the residue and atomic levels. The refined representations are then decoded into 3D coordinates via a diffusion model. During this iterative denoising process, the model produces physically plausible structures along with confidence metrics such as predicted local distance difference test (pLDDT), a predicted aligned error (PAE), and predicted distance error (PDE).

AuroBind's fitness prediction module is trained to align predicted compound rankings with experimental binding affinities. Direct preference optimization (DPO) is used for training, with structural confidence scores serving as soft weights to guide optimization. This approach fine-tunes the model to differentiate between high- and low-fitness ligands, an essential capability for efficient virtual screening and drug development.

To further enhance efficiency in large-scale screening, we developed AuroFast—a distilled version of AuroBind focused solely on fitness prediction. AuroFast is up to 100,000× faster in prediction latency compared to AlphaFold 3 or AuroBind, enabling rapid, structure-aware screening of millions of compounds that would be computationally prohibitive using the full model **(see SI Methods 1).**

**AuroBind Training**

AuroBind's training is performed in two main phases: structure prediction and fitness optimization, each executed in two stages. In the structure prediction phase, the model initially learns local structural features through 50,000 diffusion training steps and subsequently refines the protein–ligand binding interface over an additional 20,000 diffusion training steps.

During the first structure prediction stage, AuroBind was initially trained on approximately 100,000 protein–ligand complexes curated from the PDB. Complexes with ion-only interfaces or low-quality contacts were filtered out. Following AlphaFold3's data processing protocols, we selected structures released before September 30, 2021, with a reported resolution of ≤9 Å and containing ≤300 polymer chains. We also removed hydrogen atoms, chains with all unknown residues or less than 4 resolved residues, unstable ligand fragments (e.g., leaving groups), and chains with severe clashes (defined as >30% of atoms within 1.7 Å of atoms in another chain). For assemblies with more than 20 chains, we selected 20 chains based on their spatial proximity to a randomly chosen interface token.

In the second structure prediction stage, we employed self-distillation. After training on PDB[49] complexes, AuroBind was used to predict protein–ligand structures on the BindingDB dataset, which contains nearly 500,000 complexes, after excluding pairs with $IC_{50}$ values below 100 nM. We then retained pairs with ipTM > 0.8, ligand_pTM > 0.5, and protein_pLDDT > 70, while removing those with homologs in the PoseBuster datasets exhibiting more than 60% sequence similarity. This filtering yielded approximately 230,000 self-distilled protein–ligand complexes. In subsequent training, samples were drawn evenly from the original PDB dataset and the self-distilled set, with a batch size of 192 for 20,000 steps.

For fitness optimization, the first stage involved fine-tuning AuroBind on approximately 1.27 million protein–ligand pairs collected from chemogenomic databases including ChEMBL[23], PubChem[24], and ExCAPE-DB[25]. Experimentally measured binding constants served as supervision labels during this stage. In the second fitness training stage, we fine-tuned the model on a large and diverse set of protein–ligand pairs with associated binding data, using direct preference optimization (DPO) to enhance the model's ability to rank relative binding fitness. Overall training was conducted over 30 days using 96 NVIDIA H800 GPUs with Bfloat16 mixed-precision optimization, ensuring both efficiency and high model performance.

**Preference-Driven Fitness Prediction with AuroBind**

We implemented a two-stage training for fitness prediction.

**Stage 1: Supervised Fine-tuning (SFT):**

A fitness prediction head is applied to the PairFormer outputs, operating on multi-scale interaction matrices $H_{pair} \in \mathbb{R}^{N \times N \times D}$ capturing pairwise residue-ligand interactions and **Dense scoring layers** $f_{aff}(H_{pair}) = W_2 \sigma(W_1 H_{pair} + b_1) + b_2$. Trained on 1.27 million protein-ligand complexes with experimental fitness measurements, we minimize:

$$\mathcal{L}_{SFT} = \frac{1}{B} \sum_{i=1}^{B} \left( \frac{y_{pred}^{(i)} - y_{true}^{(i)}}{\sigma_{exp}^{(i)}} \right)^2 + \lambda \parallel \nabla_x y_{pred} \parallel_2,$$

where the gradient penalty term enforces smooth fitness landscape.

**Stage 2: Direct Preference Optimization (DPO):**

From Stage 1 predictions, we construct a 0.2 million complex subset filtered by confidence (confidence score > 0.9). For each complex $c_j$, we generate **Predicted fitness ranking** $R_{pred}^{(j)} = argsort(y_{pred}^{(j)})$ and Experimental fitness ranking: $R_{true}^{(j)} = argsort(y_{true}^{(j)})$. And The loss implements a Plackett-Luce ranking model, it retains the core philosophy of DPO—optimizing model outputs to reflect experimentally derived compound rankings rather than absolute fitness values:

$$\mathcal{L}_{DPO} = -\frac{1}{B} \sum_{j=1}^{B} log P(R_{true}^{(j)} | R_{pred}^{(j)}),$$

$$P(R_{true} | R_{pred}) = \prod_{k=1}^{K} \frac{exp(y_{pred}^{(r_k)}/\tau)}{\sum_{m=k}^{K} exp(y_{pred}^{(r_m)}/\tau)},$$

with $r_k$ denoting the $k$-th ranked compound in $R_{true}$, and temperature $\tau = 0.1$ controlling ranking sharpness. This preference-driven training encourages the model to prioritize rank consistency over absolute prediction error, improving its ability to distinguish compounds with subtle fitness differences. While our method remains fully differentiable and compatible with standard supervised pipelines, it aligns conceptually with the core goals of DPO and is particularly effective in structure-based screening applications where relative prioritization is essential.

**AuroFast Training**

AuroFast is a lightweight student model distilled from AuroBind for initial round high-throughput virtual screening. AuroBind takes the pair representations of protein-ligand complexes and predicts the fitness score without explicitly predicting the complex structures.

For a given protein target, compounds in the screening library are first clustered based on their ECFP4 fingerprints (radius = 2, 1024 bits). Within each cluster, a centroid compound $c_{cluster}$ is selected based on maximum average Tanimoto similarity to other members (intra-cluster similarity > 0.6). During training, the structural embedding (referred to as the trunk embedding) of the corresponding $c_{cluster}$ –protein complex, as predicted by AuroBind, is used as a prior. AuroFast takes two types of input features:

(1) Base features, including the compound's ECFP4 fingerprint and the protein's ESM-2 embedding;

(2) Augmented features, comprising the structural embedding of the nearest centroid complex from AuroBind.

The model is trained using a multi-task loss to jointly approximate structural features and binding fitness:

$$\mathcal{L}_{fast} = \alpha \parallel H_{pred} - H_{main} \parallel_2 + \beta |y_{pred} - y_{true}|,$$

where $H_{pred}$ denotes the hidden representation of predicted trunk embedding, $H_{main}$ is the teacher model's the hidden representation of trunk embedding, and $y_{pred}$ denotes the fitness estimate. At inference, novel compounds are encoded by retrieving their nearest $c_{cluster}$, the number of clusters is adaptively determined based on the dataset size—for instance, one cluster center per 100,000 compounds, generating hybrid features through concatenation of ECFP4, ESM-2, and $c_{cluster}$'s structural prior, and predicting fitness via a lightweight Transformer. This architecture accelerates screening to 830 compounds/sec per H800 while maintaining comparable accuracy (Fig. 2d, Fig. 2e), as structural priors from $c_{cluster}$ embeddings enable local analogical reasoning without full complex prediction from AuroBind. The training protocol involved the same database as AuroBind, optimized over 24 hours on one H800 GPUs.

**Virtual Screening Pipeline**

To showcase the performance of our virtual-screening pipeline, we carried out a large-scale screen against ~30 million molecules sourced from the we conducted a large-scale virtual screen targeting approximately 30 million molecules from sourced from the ZINC, commercial off-the-shelf drug library[44] and MedChemExpres (MCE) libraries[45], acknowledging that some overlap exists between these providers. The entire library was first evaluated with AuroFast with ten NVIDIA H800 GPUs; for each protein target this rapid single-pass predictor finished in ~6 h and returned the top-ranked 10,000 candidates (**Supplementary Fig. 6**). These 10,000 molecules were subsequently re-scored by AuroBind, which generates full protein–ligand complex models and refined fitness scores (**Supplementary Fig. 7**); one run per target completed in 24 h on a single H800 GPU on average (runtime scales with protein length), and the 500 best-scoring compounds were forwarded to cheminformatic post-processing.

Each of the 500 compounds was required to satisfy all of the following permissive property windows: molecular weight ≥ 200 Da, cLogP ≤ 6, hydrogen-bond donors ≤ 4, hydrogen-bond acceptors ≤ 10, and ESOL-predicted log S ≥ –9; structures failing basic valence checks were discarded automatically. To ensure structural novelty, compounds with an ECFP4 (radius = 2) Tanimoto similarity > 0.60 to any reported active for the same target in ChEMBL were removed. As our workflow does not rely on predefined binding pockets, no additional pose-based filters were imposed; instead, AuroBind-predicted complexes were inspected manually for plausible interactions, absence of severe clashes, and reasonable ligand strain. We selected ~50 compounds per target for wet-lab validation, and due to differences in compound availability and solubility, an average of 30–50 compounds per target were ultimately validated.

**Experimental-validation overview**

Each target-specific shortlist (30–50 compounds) was screened at a fixed concentration of 10 µM in biological triplicate (n = 3). For the ATP-competitive kinases GSK3α/β, CDK2, HER3 and TrkB, reactions were assembled in 384-well plates and read either with the ADP-Glo® luminescent kit (Promega) or with LANCE™/LanthaScreen™ TR-FRET tracer-displacement formats, exactly as described in Supplementary Methods 2.4. Recombinant CDK2 and bovine cyclin A2 were co-expressed in E. coli with yeast CAK, purified by Ni-affinity followed by size-exclusion chromatography, concentrated, aliquoted and stored at −80 °C; the other kinase preparations and all tracer reagents were obtained commercially. Reaction progress was recorded on Synergy Neo, PHERAstar FSX or EnVision 2104 readers, and signals were normalised to vehicle controls containing 1 % (v/v) DMSO.

Class-I HDAC3 activity was quantified in OptiPlate-384 wells with the fluorogenic Ac-LGK(Ac)-AMC substrate and trypsin developer; fluorescence ($\lambda\_ex$ 355 nm / $\lambda\_em$ 460 nm) was monitored continuously for 30 min at 25 °C. GPCR signalling assays were performed on a FLIPR Tetra high-throughput imager. CCR4 or HEK293-mGluR5 cells (20 000 cells well$^{-1}$) were loaded with Fluo-4 or Fluo-8 calcium indicators, pre-incubated with test compounds for 15–50 min, and stimulated with CCL17 or l-glutamate; peak fluorescence responses were referenced to high- and low-control wells containing vehicle or a known antagonist. GPR151 activation was tracked by BRET2 using an Rluc8/GFP2 G-protein biosensor, whereas GPR160 signalling was monitored with the GloSensor-22F cAMP reporter according to the manufacturer's instructions.

Compounds eliciting ≥50 % modulation relative to the appropriate control advanced to an 8–10-point, three-fold serial dilution series. Concentration–response data were fitted with a four-parameter logistic model in GraphPad Prism or XLFit, yielding IC$_{50}$ values for confirmed hits. Catalog numbers, buffer compositions and detailed plate layouts for every assay are provided in Supplementary Methods 2.

**Acknowledgements**

This study has been supported by the National Key R&D Program of China [2023YFF1205103 to J.Z.],the National Natural Science Foundation of China [62041209 to S.Z., 82373881 to J.D., 81925034 to J.Z, 22237005 to J.Z.], the innovative research team of high-level local universities in Shanghai [SHSMUZDCX20212700 to J.Z.], the Guangdong Basic and Applied Basic Research Foundation [2023A1515012616], the Young Elite Scientists Sponsorship Program by CAST [2022QNRC001 to J.D., 2023QNRC001 to S.Z.], the Shanghai Sailing Program [23YF1456800 to J.D.], the Youth Innovation Promotion Association of Chinese Academy of 896 Sciences [to J.D.], The Talent Plan of Shanghai Branch, Chinese Academy of Sciences, [to J.D]. the Science and Technology Commission of Shanghai Municipality [24510714300 to S.Z.] and the project from Smart Medical Innovation Technology Center-GDUT [ZYZX24-011 to S.Z.]. S. Z. acknowledges funding from the Asian Young Scientist Fellowship. S.S. acknowledges funding from the Shanghai Artificial Intelligence Laboratory.

## Conflict of Interest

The authors declare no competing interests.

## Author Contributions

S.Z., J. Zhang, S.S. and J.D. conceived and supervised the project. Z. Z., J.R., W.B. and J. Zhong conceived the project. D.W. and H.F. performed the GloSensor cAMP and BRET assays. J. Zhong and S.B.N. conducted kinase activity assays. L.Q. train the structure prediction model. W.B. train the structure prediction model and fitness prediction model. R.M and Z. Z train the fast fitness prediction model. X.C.S. contributed to calcium flux assays. H.Y. and R.L. contributed to computational virtual screening analyses. J. Z. also contributed to assays of other targets and wrote the assay methods. Z. Z. and J. R. implemented the virtual-screening pipeline, while L.W. and J. Z. curated the screening library and guided the in-silico workflow. W.O. and X.M. guide the experiments and analyze the results. Z.Z., J.R., R.M. and J.W. collated and interpreted the experimental results. S.Z., Z.Z., W.H., O.Z., J.X.C., W.B., wrote the manuscript with input from all authors.

## Data and Materials Availability:

A lightweight version of the code and pretrained weights is available at https://github.com/GENTEL-lab/AuroBind.

## Supplementary Materials

Supplementary Figures S1 to S8
Supplementary Tables 1 to 4
Supplementary Text
Materials and Methods

# Supplementary Materials for

# Fitness aligned structural modeling enables scalable virtual screening with AuroBind

## 1. Training pipeline

AuroBind follows similar data processing procedures as AlphaFold 3 (AF3) [21]. Fitness screening for proteins and ligands eliminates the need for processing DNA/RNA data. Here we describe the standard operations for data preparation.

### 1.1.1 Overview

AuroBind follows AF3 for parsing, cropping and featurization, with the DNA/RNA-specific components removed.

• **Parsing**: For mmCIF format inputs, we parse the atom sites and basic metadata including resolution, release data, method, bioassembly, chemical components, chain names, sequences and covalent bonds.

• **MSA:** multiple-sequence alignments are generated by ColabFold (MMseqs2 + HHblits).

• **Tokenization:** For proteins and ligands are tokenized standard amino acid residues and atoms, respectively. For proteins, $C_\alpha$ is defined as the token center atom for each amino acids. For ligands, the only atom of each token is the token center atom.

• **Cropping:**

1) **Contiguous cropping.** Contiguous sequences of polymer residues and ligand atoms are selected (AlphaFold-Multimer/AF3).
2) **Spatial cropping.** Polymer residues and ligand atoms are selected to those within close spatial distance of a reference atom randomly selected from the center atoms.
3) **Spatial interface cropping.** The reference atom is replaced with interface atom which is randomly selected from the center atoms with distance under 15Å to another chain's token center atom.

    Ligand atoms are ensured always retained.

• **Featurization:**

1) **Token feature.** The per-token feature includes position indexes, chain identifiers and masks.

2) **Reference feature.** Features are obtained from the reference conformation of a specified residue or ligand. The generation of this conformer employs RDKit, utilizing an input CCD code or SMILES string. Missing coordinates are set to zeros.

3) **MSA feature.** Features derived from MSA, deletion matrix and profile.

4) **Bond feature.** Bond information including expected locations of polymer-ligand bonds, intra-/inter-ligand bonds, etc.

**Supplementary Table 1** | Model Input Features List

| Feature Type | Description |
| --- | --- |
| residue_index | Residue number in the token's original input chain. |
| token_index | Token number. Increases monotonically; does not restart at 1 for new chains. |
| asym_id | Unique integer for each distinct chain. |
| entity_id | Unique integer for each distinct sequence. |
| sym_id | Unique integer within chains of this sequence. E.g. if chains A, B and C share a sequence but D does not, their sym_ids would be [0, 1, 2, 0]. |
| restype | One-hot encoding of the sequence. 32 possible values: 20 amino acids + unknown, 4 RNA nucleotides + unknown, 4 DNA nucleotides + unknown, and gap. Ligands represented as "unknown amino acid". |
| is_protein / rna / dna / ligand | 4 masks indicating the molecule type of a particular token. |
| ref_pos | Atom positions in the reference conformer, with a random rotation and translation applied. Atom positions are given in Å. |
| ref_mask | Mask indicating which atom slots are used in the reference conformer. |
| ref_atom_name_chars | One-hot encoding of the unique atom names in the reference conformer. Each character is encoded as ord(c) − 32, and names are padded to length 4. |
| ref_space_uid | Numerical encoding of the chain id and residue index associated with this reference conformer. Each (chain id, residue index) tuple is assigned an integer on first appearance. |
| ref_element | One-hot encoding of the element atomic number for each atom in the reference conformer, up to atomic number 128. |
| ref_charge | Charge for each atom in the reference conformer. |
| msa | One-hot encoding of the processed MSA, using the same classes as restype. |
| has_deletion | Binary feature indicating if there is a deletion to the left of each position in the MSA. |
| deletion_value | Raw deletion counts (the number of deletions to the left of each MSA position) are transformed to [0, 1] using $(2/\pi)$ arctan(d/3). |
| profile | Distribution across restypes in the main MSA. Computed before MSA processing. |
| deletion_mean | Mean number of deletions at each position in the main MSA. Computed before MSA processing. |
| token_bonds | A 2D matrix indicating if there is a bond between any atom in token i and token j, restricted to just polymer-ligand and |

| | ligand-ligand bonds and bonds less than 2.4 Å during training. |
|---|---|

- **Training Set Clustering：**

To reduce bias in the training and evaluation sets, clustering was performed on protein chains and small-molecule ligands as follows:

1) Protein chains were clustered at 40% sequence identity.

2) Small molecules were clustered based on chemical component dictionary (CCD) identity, meaning only chemically identical molecules were grouped into the same cluster.

No additional interface-based clustering was applied, as the dataset consists solely of protein–small molecule complexes without other interface types.

### 1.1.2 Training data

**Structure prediction training data:**

During the first structure training stage, AuroBind was trained on nearly 100,000 protein–ligand complexes curated from the PDB[22]. Complexes with ion-only interfaces or low-quality contacts were filtered out. Following AlphaFold3's data processing protocols, we selected structures released before September 30, 2021, with a reported resolution of ≤9 Å, containing ≤300 polymer chains. We also removed hydrogen atoms, polymer chains with all unknown residues or less than 4 resolved residues, unstable ligand fragments (e.g., leaving groups), and chains with severe clashes (defined as >30% of atoms within 1.7 Å of atoms in another chain). For assemblies with more than 20 chains, we selected 20 chains based on their spatial proximity to a randomly chosen interface token.

In the second structure training stage, we employed self-distillation.

**Self-distilled prediction training data:**

After training on PDB complexes, AuroBind was used to predict protein–ligand structures on the BindingDBdataset, which contains nearly 500,000 complexes, after excluding pairs with IC50 values below 100 nM. We then retained pairs with ipTM > 0.8, ligand_pTM > 0.5, and protein_pLDDT > 70, while removing those with homologs in the PoseBuster[42] datasets exhibiting more than 60% sequence similarity. This filtering yielded 230,000 self-distilled protein–ligand complexes, and in subsequent training, the training samples were evenly divided—50,000 samples per epoch, with half coming from PDB data and half from the self-distilled set.

**Fitness Finetuning data:**

We curated a large-scale chemogenomic dataset, AuroDB, by integrating data from ChEMBL30[23] and PubChem[24], following the data cleaning strategy outlined in ExCAPE-

DB[25]. Specifically, 58,235 and 92,147 single-target concentration-response (CR) assays were selected from PubChem and ChEMBL, respectively. Active compounds with dose–response values ≤10 μM were retained, while inactive compounds from CR assays and screening assays were included.

The dataset originally comprised 72,161,011 structure–activity relationship (SAR) data points, each assigned a log-transformed activity value (pXC50). To improve data quality, we removed compounds with infrequent appearances within the dataset and excluded all data points with a pXC50 value of 3.101, which were expert-defined negative samples with low confidence. These steps resulted in a more balanced and high-confidence dataset.

The final benchmark dataset contained 3,636,740 chemogenomic data points, corresponding to 2,136,238 unique compounds. To further optimize computational efficiency, we excluded target proteins with sequence lengths exceeding 1000. Additionally, we applied down-sampling based on the distribution of fitness values, resulting in a refined dataset of 1,273,781 data points, covering 1376 target proteins and 492,447 unique compounds. This curated dataset was ultimately used for fine-tuning AuroBind. Experimentally measured binding constants served as supervision labels during this stage. In the second fitness training stage, we selected protein–ligand pairs that exhibited predicted structure confidence scores greater than 0.5 and employed these for directed preference optimization (DPO) training.

**1.2 Model Framework**

**1.2.1 Fitness head**

The structural model's 48-layer PairFormer parameters are frozen and excluded from training. A new 4-layer PairFormer module (with trainable parameters) is introduced to further process features output by the frozen model, producing single and pair features. Single features encompass all tokens, while pair features focus on protein-ligand and ligand-specific interactions, excluding protein-protein pair features. These features are fed into a fitness prediction head (primarily an MLP structure) to predict fitness values for each residue (for proteins) or atom (for ligands). The total fitness per token is the sum of both contributions.

To address the disparity in sequence lengths between proteins and ligands, the loss function weights are adjusted: the total weight for the small molecule part is set to twice that of the protein part. Within each modality, weights are equal across residues or atoms, and all weights are subsequently normalized. The overall fitness prediction is computed as the weighted sum of each token's fitness value and its normalized weight. The algorithm process is as follows.

```
Algorithm 1 LigandAffinityHead
        def LigandAffinityModule(s, z, single_mask, pair_mask,
            is_ligand, num_protein, num_ligand, T):
        # Pairwise attention encoding
 1:     s, z ← PairformerStack(s, z, single_mask, pair_mask)
        # Add sequence residual
 2:     s ← s + s_update
        # Add pair residual
 3:     z ← z + z_update
        # Token-level contribution
 4:     s_affinity ← σ(W_s·LN(s)) ⊙ MLP(LN(s))
        # Pair-level aggregation
 5:     z_affinity ← ∑(σ(W_z·LN(z)) ⊙ MLP(LN(z)), dim = −2)
                    / ∑(head_pair_mask)
        # Combine affinities
 6:     batch_affinity ← s_affinity + z_affinity
        # Ligand-specific weighting
 7:     ratio ← ligand_weight_ratio × (num_protein/num_ligand)
 8:     affinity_weight ← is_ligand × ratio + (1 − is_ligand)
        # Normalize weights and compute affinity
 9:     norm_weight ← softmax(affinity_weight/T)
10:     affinity ← ∑(batch_affinity ⊙ norm_weight, dim = −2)
11:     return affinity
```

### 1.2.2 Confidence head

The confidence head leverages the same PairFormer outputs as the fitness head, processed through additional MLP layers to generate these metrics. Training enables the model to predict several outputs, including three confidence metrics: the predicted local distance difference test (pLDDT, measuring per-atom confidence), the predicted aligned error (PAE), and the predicted distance error (PDE).

• **pLDDT.** Predict LDDT that takes into account distances from all atoms to polymer residues. For a ligand atom the confidence only takes into account interactions between the ligand atom and proteins.

• **PAE.** The estimate of the error of one token when aligned according to the frame of another helps determine confidence of interfaces or specific interactions between atoms.

• **PDE.** Predict the error in absolute distances between atoms.

### 1.3 Finetuning and Optimization

AuroBind's training is performed in two main phases: structure prediction and fitness optimization, each executed in two stages. In the structure prediction phase, the model initially learns structural features through 50,000 steps and subsequently refines the protein–ligand binding interface over an additional 20,000 steps.

- **Stage 1 (structure):** 10 epochs self-distillation, crop 768, dynamic sampling favoring interface-rich crops.

- **Stage 2 (fitness):** 20 epochs on ExCAPE-DB[25], identical crop.

- **Optimization:** Adam ($\beta 1 = 0.9$, $\beta 2 = 0.95$, $\varepsilon = 1e\text{-}8$). Base LR = 1.8e-3 (Stage 1) or 2.0e-4 (Stage 2); warm-up 1 k steps, decay ×0.95 every 50 k steps. Batch size 192 (structure) / 96 (fitness). Gradient clipping at norm 10; dropout 0.1.

- **Validation:** 4 % complexes held out by scaffold split (Bemis-Murcko) and ≤ 40 % protein identity.

**Supplementary Table 2** | Details for AuroBind's two stage training.

| Parameter | Structural Training | Fitness Training |
|---|---|---|
| Crop Size | 384/768 | 768 |
| Learning Rate | $1.8 \times 10^{-3}$ | $2 \times 10^{-4}$ |
| LR Decay | 0.95(every 50,000) | 0.01 |
| Batch Size | 192 | 4 (per card) |

## 1.4 Data Sampling Method

To address the imbalance of samples across different protein targets, we designed a custom data sampling pipeline. First, we filtered the dataset by retaining only those proteins (UniProt ID) with at least 10 associated entries to ensure sufficient data for each group. Then, within each protein group, we discretized the continuous labels into bins using a threshold of 0.15 to cluster similar values. For each bin, we randomly selected up to 5 entries to reduce over-representation of densely sampled regions while maintaining label diversity.

To generate training batches, we implemented a custom group-aware sampler, RandomGroupBatchSampler, which ensures that each mini-batch contains samples drawn from the same protein group. This sampling strategy ensures balanced representation across protein targets while preserving intra-group label diversity and reducing label redundancy.

During the DPO training stage, we further prioritized sampling data with similar fitness values to enhance the model's ability to distinguish subtle differences in binding affinities.

## 1.5 Data and Benchmark

We evaluated AuroBind's structure prediction capabilities using the PoseBuster benchmark datasets. PoseBuster V1 comprises 428 protein–ligand complexes, while the more recent PoseBuster V2 contains 308 structurally diverse complexes released after 2021 and curated to remove any crystal contacts[42]. In keeping with AlphaFold 3's protocol[21], we ensured zero overlap between the training and validation sets. Structural accuracy was assessed by pocket aligned root-mean-square deviation (RMSD), with values below 2.0 Å indicating high-quality pose predictions, and by PoseBuster's physical plausibility suite, which checks chemical consistency, intra- and intermolecular geometry, and hydrogen bonding fidelity.

For fitness prediction, we benchmarked against the DAVIS[29] and BindingDB[30] datasets, which contain protein–ligand pairs annotated with experimentally determined dissociation constants ($K_d$). The DAVIS dataset represents a low-resource scenario with just 2,086 labeled examples, whereas BindingDB offers a larger set of 12,668 examples. We split each dataset into 70% training, 10% validation, and 20% testing, following the setting of ConPlex[57]. The definition of positive and negative samples also follows the criteria used in ConPLex. To mitigate class imbalance during model training, we up-sampled the minority class in the training set so that positives and negatives were equally represented, while preserving the natural distribution in validation and test sets. Model performance was evaluated using both the area under the precision–recall curve (AUPR) and the area under the receiver operating characteristic curve (ROC-AUC).

To assess high-throughput virtual screening performance, we turned to the Lit-PCBA[35] benchmark, a curated collection of 149 PubChem bioassays covering 15 protein targets, each associated with experimentally validated actives (IC50 < 1 µM) and physico-chemically matched decoys. In total, the benchmark comprises 7,844 actives and 407,381 decoys—a 1:52 active-to-inactive ratio—and all targets have corresponding high-resolution co-crystal structures. In a pure zero-shot setting, we applied AuroFast to the full dataset for each target and reported early enrichment using the enrichment factor at 1% (EF1%), reflecting the model's ability to prioritize true positives among the top-ranked compounds.

To obtain experimentally confirmed inactive compounds, we queried ChEMBL for each target and retained all entries that satisfied either of the following criteria: (i) a standard activity value ($IC_{50}$, $K_i$, $K_d$, $EC_{50}$ or analogous) > 20 µM; or (ii) a reported pChEMBL > 4 when the raw concentration value was unavailable. The resulting inactive pool was merged with the positive set and filtered to remove compounds appearing in both lists or with conflicting annotations. Table S3 summarizes the final counts and negative-to-positive ratios for every target used in EF1 % evaluations.

**Supplementary Table 3** | Details for definition of negative compounds in CHEMBL, using condition (Activity Unit == 'nM' and Activity Value > 20000 or pChEMBL Value < 4.5) for CHEMBL negative samples, we also included all in-house experimentally tested molecules that demonstrated less than 50% inhibition at 10 µM.

| Targets Name | Active(>50%) | Active(>90%) | Inactive |
|---|---|---|---|
| CCR4 | 4 | 1 | 76 |
| CDK2 | 6 | 2 | 733 |
| GSK-3α | 23 | 13 | 347 |
| GSK-3β | 17 | 9 | 757 |
| HDAC3 | 2 | 2 | 378 |
| HER3 | 10 | 7 | 33 |
| mGluR5 | 3 | 2 | 336 |
| TrkB | 16 | 5 | 55 |
| Total | 81 | 41 | 2752 |

## 2. Experimental Validation Protocols for Screening Targets

### 2.1 Target Description

**Glycogen synthase kinase-3 (GSK-3)**, a serine/threonine kinase with two highly homologous mammalian isoforms, GSK-3α and GSK-3β. It recognizes and phosphorylates substrates containing the conserved SXXXS(P) motif, targeting diverse cytoplasmic proteins and nuclear transcription factors. GSK-3 plays critical roles in regulating glycogen metabolism, insulin signaling, Wnt pathways and implicated in diabetes, neurodegenerative diseases and cancers. To address the challenges of isoform redundancy and off-target effects, we designed isoform-selective inhibitors to target the ATP-binding pockets of GSK-3α and GSK-3β[58-61].

**Cyclin-dependent kinase 2 (CDK2)**, a cyclin-dependent serine/threonine kinase regulating G1/S phase transition in the cell cycle, driving the phosphorylation of the Rb protein by forming complexes with Cyclin E/A, thereby releasing E2F transcription factors to promote DNA replication and cell proliferation. It is aberrantly activated in various cancers. We designed highly selective inhibitors targeting its ATP-binding pocket, leveraging subtle structural distinctions between CDK2 and CDK4/6 in the ATP-binding site to minimize off-target effects[62,63].

**Metabotropic glutamate receptor 5 (mGluR5)**, a class C G protein-coupled receptor (GPCR) widely expressed in brain regions associated with cognition and emotion, is a key modulator of excitatory neurotransmission and a promising therapeutic target in neuropsychiatric and neurodegenerative disorders. mGluR5 mediates intracellular signaling via the Gq-phospholipase C pathway and is implicated in conditions such as anxiety, depression, schizophrenia, and Fragile X syndrome. We targeted the allosteric transmembrane domain of mGluR5 using selective negative allosteric modulators (NAMs), which bind a well-characterized site distinct from the orthosteric glutamate-binding domain. Allosteric modulators, such as mavoglurant and MTEP, enable receptor subtype selectivity and show in vivo efficacy in preclinical models[64-66].

**C-C chemokine receptor type 4 (CCR4)**, a seven-transmembrane GPCR for CCL17 and CCL22, is predominantly expressed on Th2 and Treg cells, and is implicated in allergic diseases such as atopic dermatitis and asthma, and various lymphomas. We targeted the intracellular C-terminal allosteric site of CCR4, which modulates ligand-specific activation and internalization[67-70].

**Histone deacetylase 3 (HDAC3)**, a class I HDAC enzyme, plays critical roles in diverse physiological processes, including embryonic development, circadian rhythm, energy metabolism and immune modulation. Dysregulation of HDAC3 is implicated in cancer, neurodegenerative disorders, and inflammatory diseases, making it a promising therapeutic target. We designed selective inhibitors of HDAC3 by exploiting isoform-specific interactions within the active site[71-74].

**Receptor tyrosine-protein kinase (HER3)**, a member of the HER family of receptor tyrosine kinases lacking intrinsic kinase activity, contribute to oncogenic signaling via ligand-induced heterodimerization with other HER family members, particularly HER2. HER3 is altered or aberrantly expressed across a variety of tumor types and can be associated with poor clinical outcomes. Whereas anticancer agents targeting EGFR and HER2 have been approved for decades, no small molecule drug targeting HER3 has been approved. We targeted the ATP-binding site of HER3[75,76].

BDNF/NT-3 growth factors receptor (Tropomyosin receptor kinase B, TrkB, NTRK2), encoded by NTRK2, is a receptor tyrosine kinase activated by brain-derived neurotrophic factor (BDNF) and neurotrophin-4 (NT-4). Its dysregulation drives oncogenesis in cancers with NTRK2 fusions and is implicated in neurodevelopmental disorders and epilepsy. We targeted the ATP-binding pocket to find next-generation selective TrkB inhibitors[77-80].

## 2.2 Reagents

Synthetic peptides including TPX2 (residues 1-43), CDK2 substrate peptide HHASPRK, and HDAC3 substrate peptide Ac-LGK(Ac)-AMC were purchased from GL Biochem (Shanghai) Ltd. Human HDAC3/NcoR2 was purchased from BPS (San Diego, USA).

## 2.3 Protein expression and purification

### 2.3.1 Phosphorylated CDK2

To generate CDK2 phosphorylated on T160 (p-CDK2), the gene encoding human CDK2 (1-298) was cloned into pRSFDuet1 vector with an N-terminal hexahistidine tag and coexpressed with GST-tagged yeast CAK in E. coli BL21(DE3)[81,82]. The culture conditions were the same as described for Aurora A. Cell pellets were lysed in lysis buffer (50 mM Tris, pH 8.0, 500 mM NaCl, 10% glycerol, 1 mM TCEP). Lysates were centrifuged at 23,000 rpm for 30 min, and loaded onto a HisTrap FF Ni-NTA column (Cytiva), washed with lysis buffer, and eluted with elution buffer (1 × PBS, pH 7.4, 500 mM NaCl, 10% glycerol, 500 mM imidazole, 1 mM TCEP). Samples were concentrated and further purified by size exclusion chromatography (SEC) using a Superdex 75 increase 10/300 GL column (Cytiva) in SEC buffer (50 mM HEPES, pH 7.5, 150 mM NaCl, 10 mM MgCl2, 1 mM EGTA, 1 mM TCEP). Purified p-CDK2 was concentrated, aliquoted and stored at -80 °C.

### 2.3.2 Bovine cyclin A2 (residues 171-432)

The gene encoding bovine cyclin A2 (171-432) was cloned into pET28a vector with an N-terminal hexahistidine tag and expressed in E. coli BL21(DE3)[81,82]. The culture conditions were the same as described for Aurora A. Cell pellets were lysed in lysis buffer (50 mM Tris, pH 8.25, 300 mM NaCl, 100 mM MgCl2, 10% glycerol, 1 mM TCEP). Lysates were centrifuged at 23,000 rpm for 30 min, and loaded onto a HisTrap FF Ni-NTA column (Cytiva), washed with lysis buffer, and eluted with elution buffer (50 mM Tris, pH 8.25, 300 mM NaCl, 100 mM MgCl2, 500 mM imidazole, 10% glycerol, 1 mM TCEP). Samples were concentrated and further purified by size exclusion chromatography (SEC) using a Superdex 75 increase 10/300 GL column (Cytiva) in SEC buffer (50 mM Tris, pH 8.25, 100 mM MgCl2, 1 mM TCEP). Purified cyclin A2 was concentrated, aliquoted and stored at -80 °C.

## 2.4 Biochemical Assays

### 2.4.1 ADP-Glo Kinase assays

Kinase activity was measured using ADP-Glo Kinase Assay kit (Promega) in ProxiPlate 384 shallow well plates (Revvity). All kinase reactions were performed in duplicate using assay buffer containing 20 mM HEPES pH 7.4, 20 mM NaCl, 1 mM EGTA, 10 mM MgCl2, 0.02% Tween-20, 0.1 mg/mL BSA, and 50 μM DTT. In the primary screening, compounds at 1 mM were diluted 20-fold in assay buffer.

For CDK2 assay, 1 μL of diluted compound (final 10 μM) was pre-incubated with 2 μL of a mixture of CDK2 (final 5 nM) and HHASPRK (final 100 μM) at room temperature for 30 min. Reactions were initiated by adding 2 μL of a mixture of cyclin A2 (final 5 nM) and ATP (final 200 μM), and incubated at room temperature for 30 min.

Reactions were then quenched using ADP-Glo reagent according to the manufacturer's instructions. Luminescence was measured using a Synergy neo microplate reader (BioTek) with an integration time of 1 s. Data normalization was performed against vehicle control (1% DMSO). For dose-response studies, compounds were serially diluted three-fold in DMSO across seven concentration points, followed by 20-fold dilution in assay buffer to obtain a working solution at 5% DMSO. Kinase activity was determined as described above. Concentration-response curves were fitted by four-parameter nonlinear regression to generated IC50 values using GraphPad Prism 7.0 software.

### 2.4.2 Time-resolved fluorescence resonance energy transfer (TR-FRET) assays

• **HER3 inhibitor screening.**

All kinase reactions were performed using the HTRF PPI Europium detection buffer (Revvity, 61DB9RDF). Test compounds were initially dissolved in DMSO to prepare 50 mM stock solutions, then transferred (150 nL per well) into a 384-well microplate (Greiner, 784075) using an Echo 655 liquid handler (Bechman Coulter). A 5 μL of HER3 kinase (1 nM; SignalChem, E28-35G-20) was dispensed into each well of the assay plate. The plate was centrifuged at 1000 rpm for 1 min, and incubated at 25°C for 10 min. Then, 5 μL of Tracer 178 (10 nM; Invitrogen, PV5593) and 5 μL of anti-GST mAb Eu-conjugate (100×; Revvity, 61GSTKLA) were added to each well of the plate, and incubated at 25°C for 1 h. Fluorescence signals were measured at 620 nm and 665 nm using a BMG PHERAstar FSX microplate reader. Concentration-response curves were fitted by four-parameter nonlinear regression to generated IC50 values using GraphPad Prism 7.0 software.

- **GSK3α and GSK3β inhibitor screening.**

Kinase activity of GSK3α and GSK3β were evaluated using LANCE Ultra TR-FRET assay. Test compounds were diluted to 100× the final assay concentration and transferred (200 nL per well) to 384-well assay plates using an Echo 650 liquid handler (Bechman Coulter). Then, 10 μL of GSK3α (final 1 nM; Carna, 04-140) or GSK3β (final 1 nM; Carna, 04-141) kinase solution was added to each well of the assay plate, except for control wells with only 10 μL of 1× kinase buffer (50 mM HEPES pH 7.5, 10 mM $MgCl_2$, 2 mM DTT ,0.01% BSA and 0.01% Triton X-100), and pre-incubated at room temperature for 10 min. Reactions were initiated by adding 10 μL of a mixture of ATP (final 10 μM for GSK3α and 7.6 μM for GSK3β) and ULight™-4E-BP1 peptide substrate (final 0.015 μM; PerkinElmer, TRF0128-M) and incubated at room temperature for 60 min. The reaction was stopped by adding 20 μL of detection solution containing 10× LANCE detection buffer, 20 mM EDTA (Gibco, 15575-038) and Eu-anti-phospho-4E-BP1 antibody (final 0.05 nM; PerkinElmer, TRF0216-M), and incubated at room temperature for another 60 min. TR-FRET signals were measured on an EnVision 2104 multilabel reader (PerkinElmer) using emission wavelengths of 615 nm (reference) and 665 nm (signal). Signal ratios (665 nm/615 nm) were normalized to vehicle control (1% DMSO). Dose-response curves were generated using 10-point three-fold serial dilutions. Concentration-response curves were fitted by four-parameter nonlinear regression to generated IC50 values in Microsoft Excel using the XLFit add-in (version5.4.0.8; IDBS).

- **TRK-B inhibitor screening**

Kinase activity of TRK-B were evaluated using the LanthaScreen™ TR-FRET assay. Test compounds were diluted to 100× the final assay concentration and transferred (200 nL per well) to 384-well assay plates using an Echo 650 liquid handler (Bechman Coulter). Then, 10 μL of TRK-B (final 0.2 nM; Carna, 08-187) kinase solution was added to each well of the assay plate, except for control wells with only 10 μL of 1× kinase buffer (50 mM HEPES pH 7.5, 10 mM $MgCl_2$, 2 mM DTT ,0.01% BSA and 0.01% Triton X-100), and pre-incubated at room temperature for 10 min. Reactions were initiated by adding 10 μL of a mixture of ATP (final 5 μM for TRK-B) and Fluorescein-Poly GT (final 0.02 μM; Invitrogen, PV3610) and incubated at room temperature for 60 min. The reaction was stopped by the addition of 20 μL of detection solution containing 10× LanthaScreenTM detection buffer, 20 mM EDTA (Gibco, 15575-038) and Tb-anti-phosphotyrosine (PY20) antibody (final 1 nM; Invitrogen, PY3529), and incubated at room temperature for another 60 min. TR-FRET signals were measured on an EnVision 2104 multilabel reader (PerkinElmer) using excitation wavelength at 340 nm and dual emission wavelengths at 495 nm (reference) and 520 nm (signal). Signal ratios (520 nm/495 nm) were normalized to vehicle control (1% DMSO). Dose-response curves were generated using 10-point three-fold serial dilutions. Concentration-response curves were fitted by four-parameter nonlinear regression to generated IC50 values in Microsoft Excel using the XLFit add-in (version5.4.0.8; IDBS).

### 2.4.3 Calcium flux assays

• **CCR4 inhibitor assay**

CCR4 cells were seeded at a density of 20000 cells/well in 20 μL medium into 384-well cell culture plates and incubated overnight at 37°C, 5% CO2 incubator. For primary screening, 900 nL of test compounds were acoustically transferred into 384-well compound plates using an Echo 555 liquid handler (Labcyte). Then, 30 μL of assay buffer was dispensed into each well. A 20 μL of 2× Fluo-4 DirectTM No-wash loading buffer (Invitrogen) was gently added to the cell culture plate, followed by 10 μL of compounds solutions from the compound plate to the cell plate. The plate was incubated for 50 min at 37°C, 5% CO2 and equilibrated for additional 10 min at room temperature. To initiate calcium flux, 10 μL of CCL17 (final 25 nM) was added to each well. Fluorescence signals were immediately monitored at Ex/Em 494 nm/516 nm using the FLIPR Tetra high-throughput cellular screening system (Molecular Devices). Data normalization was performed against high control (0.5% DMSO) and low control (10 μM AZD-2098). For dose-response studies, compounds were serially diluted three-fold in DMSO for ten concentration points using the Echo 555 system. Concentration-response curves were fitted by four-parameter nonlinear regression to generated IC50 values using GraphPad Prism 5.0 software.

• **mGluR5 inhibitor assay**

HEK293 cells expressing human mGluR5 (HD Bioscience) were plated in black-walled 384-well imaging plates (PerkinElmer, 6007290) at a density of 20000 cells/well in complete growth medium. In the primary screening, test compounds were diluted 44.4-fold from 4 mM DMSO stocks into assay buffer (20 mM HEPES pH 7.4 and 0.1% BSA) to achieve working solutions containing 2.25% DMSO. On the day of the experiment, the medium was aspirated and replaced with 40 μL of Fluo-8 AM working solution (final 8 μM; AAT Bioquest, 21080). The plates were incubated at 37°C, 5% CO2 for 30 min. A 5 μL aliquot of each diluted compound was added to dye-loaded cells and pre-incubated at room temperature for 15 min. Calcium flux was triggered by the addition of 10 μL L-glutamate (final 2.31 μM). Fluorescence signals were immediately monitored at Ex/Em 490 nm/525 nm using the FLIPR Tetra high-throughput cellular screening system (Molecular Devices) with a negative allosteric modulator (NAM) assay format. Data normalization was performed against the vehicle control (0.25% DMSO). For dose-response studies, compounds were serially diluted three-fold in DMSO for eleven concentration points using the Echo 655 system. Concentration-response curves were fitted by four-parameter nonlinear regression to generated IC50 values in Microsoft Excel using the XLFit add-in (version5.5; IDBS).

### 2.4.4 Fluorescence-based HDAC3 assay

Enzyme activity was measured using a fluorescence-based assay in OptiPlate 384 well plates (Revvity, 6007279). All reactions were performed in duplicate using assay buffer containing 50mM Tris pH 7.5, 50mM NaCl and 0.01% Tween-20. Test compounds were dissolved in DMSO to 50 mM stock solutions and diluted to 100× the final assay concentrations. For enzymatic inhibition studies, 250 nL of each diluted compound was transferred to the assay plate, followed by 15 μL of HDAC3 (final 7 nM), and pre-incubated at room temperature for 15 min. Reactions were initiated by adding 10 μL of a mixture containing Ac-LGK(Ac)-AMC peptide (final 5 μM) and trypsin (final 0.05 μM; Sigma-Aldrich). Fluorescence kinetics were monitored for 30 min at 25°C using a Paradigm microplate reader (Molecular Devices) with excitation/emission wavelengths of 355 nm and 460 nm, respectively. Data normalization was performed against vehicle control (1% DMSO). For dose-response studies, compounds were serially diluted three-fold in DMSO across ten concentration points to obtain a working solution. Enzyme activity was determined as described above. Concentration-response curves were fitted by four-parameter nonlinear regression to generated IC50 values in Microsoft Excel using the XLFit add-in (version5.4.0.8; IDBS).

### 2.4.5 BRET2 assay

The full-length GPR151 were cloned into pcDNA6.0 vector (Invitrogen) with a FLAG tag at its N-terminus. Human GαoA-Rluc8, Gβ3 and Gγ8-GFP2 used TRUPATH platform. HEK-293T cells (ATCC, # CRL-11268) were cultured in Dulbecco's Modified Eagle Medium (Gibco) supplemented with 10% (w/v) fetal bovine serum. Cells were maintained at 37 ˚C in a 5% CO2 incubator with 3.5×10⁵/mL cells per well in a 6-well plate. Cells were grown overnight and then transfected with 0.75 μg GPR151, 0.75 μg GαOA-Rluc8, 0.75 μg Gβ and 0.75 μg Gγ8-GFP2 constructs by YEASEN® Hieff Trans Liposomal Transfection Reagent in each well for 48 h. Cells were harvested and re-suspended in Tyrode's solution buffer at a density of 5×10⁵ cells/mL. The cell suspension was seeded in a 96-well plate at a volume of 70 μL per well, 20 μL Tyrode's solution buffer containing 10μM of ligands (or 10-fold serial dilution of ligand), and another 10 μL the Nano-Glo® Live Cell Substrate (CTZ-400a, 1:100 dilutions, YEASEN) diluted in the detection buffer. The luminescence signal was measured with a BioTEK plate reader at room temperature. The BRET signal is determined by calculating the ratio of the light emitted by the RLuc8-coelenterazine 400a (410nm) and GFP2 (515nm). The average baseline value (basal BRET ratio) recorded prior to agonist stimulation was subtracted from the experimental BRET signal values to obtain the resulting difference (BRET ratio). All concentration–response curves were fit to a three-parameter logistic equation in Prism (Graphpad Software). BRET concentration–response curves were analyzed as either raw net BRET2 (fit Emax-fit Baseline) for each experiment. Data were normalized to the ligands induced G protein recruitment to GPR151.

### 2.4.6 GloSensor cAMP inhibition assay

The full-length GPR160 was fused with HA signal peptide and FLAG epitope in the N terminus, which was cloned into the pcDNA6.0 vector. Before transfection, HEK-293T cells were plated onto six-well plates with density of 3×10⁵ cells per mL. After 24h, cells were transfected with 1.5 μg of receptor and 1.5 μg of GloSensor-22F (Promega). After 48h, cells were starved by Hank's balanced salt solution for 30 min. Subsequently, the cells were digested and transferred onto 384-well plates with 20 μL of CO2-independent media suspension containing 2% GloSensor cAMP Reagent (Promega) with a density of 4×10⁵ cells per mL. After 1h incubation, 10 μL of 10 μM ligands (or 10-fold serial dilution of ligand) were added and incubated for 10 min at room temperature, the ligand buffer contains Forskolin (MCE) with a final concentration of 1μM. All luminescence signals were measured using the EnVision multi-plate reader in accordance with the manufacturer's instructions. Data were normalized to the ligands induced cAMP inhibition by wild-type receptor.

### 2.4.7 Effects of GPR160 inhibitors on cell proliferation

PC3 (3,000 cells/well) and HepG2 (5,000 cells/well) cells were seeded into 96-well plates. The starting point of the assay was defined after cell attachment. Cells were treated with GPR160-C17 at concentrations of 0 μM, 5 μM, 10 μM, and 50 μM for 0, 24, 48, and 72 hours. At the indicated time points, cells were fixed with 100 μl of 10% (w/v) trichloroacetic acid (TCA) per well. Fixed cells were stained with 4 mg/ml sulforhodamine B (SRB) in 1% (v/v) acetic acid, and the bound dye was solubilized with 200 μl of 10 mM Tris-HCl. Absorbance was measured at 560 nm using a microplate reader.

## 3. Supplementary Results

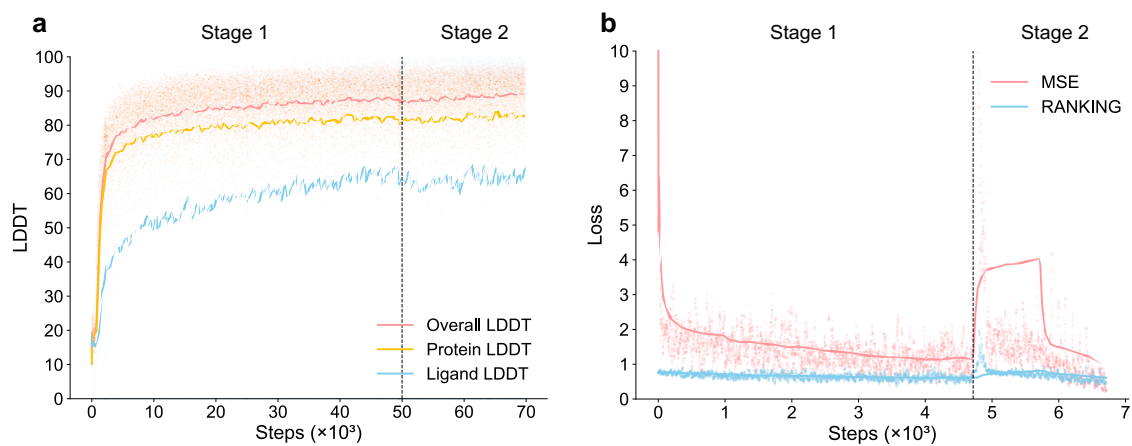

**Supplementary Fig. 1 | Accuracy across training.**
**a,** Structure-prediction accuracy during two-stage training. Stage 1 (left of the dashed line) corresponds to optimization on experimentally determined protein–ligand complexes, whereas Stage 2 fine-tunes the network on a self-distilled corpus of high-confidence and high-fitness predictions. Running means (solid lines) and individual mini-batch scores (dots) are shown for overall LDDT (pink), protein LDDT (gold) and ligand LDDT (blue).
**b,** Supervised optimization of the fitness. In Stage 1 the model is trained with a mean-squared-error (MSE) objective (red) while a ranking loss (blue) is monitored but not optimized. In Stage 2 the MSE objective is replaced by directed preference optimization (DPO), producing a transient spike in MSE that quickly subsides as both losses converge to lower values.

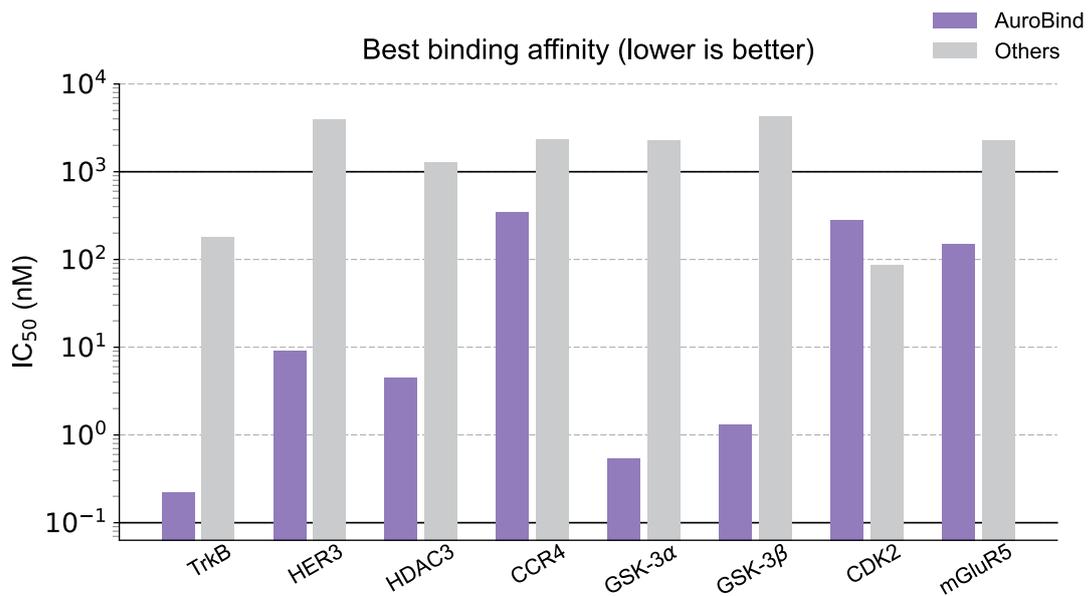

**Supplementary Fig. 2 | AuroBind delivers more potent binders than previous virtual-screening workflows across eight therapeutic targets.** Bars represent the lowest experimentally measured IC$_{50}$ (nM; lower values indicate higher fitness) obtained with AuroBind (purple) versus the best compounds reported by earlier in-silico campaigns ("Others", grey). Across all targets, AuroBind matches or surpasses the fitness of both the leading unoptimized molecules and those improved through multiple rounds of wet-lab optimization.

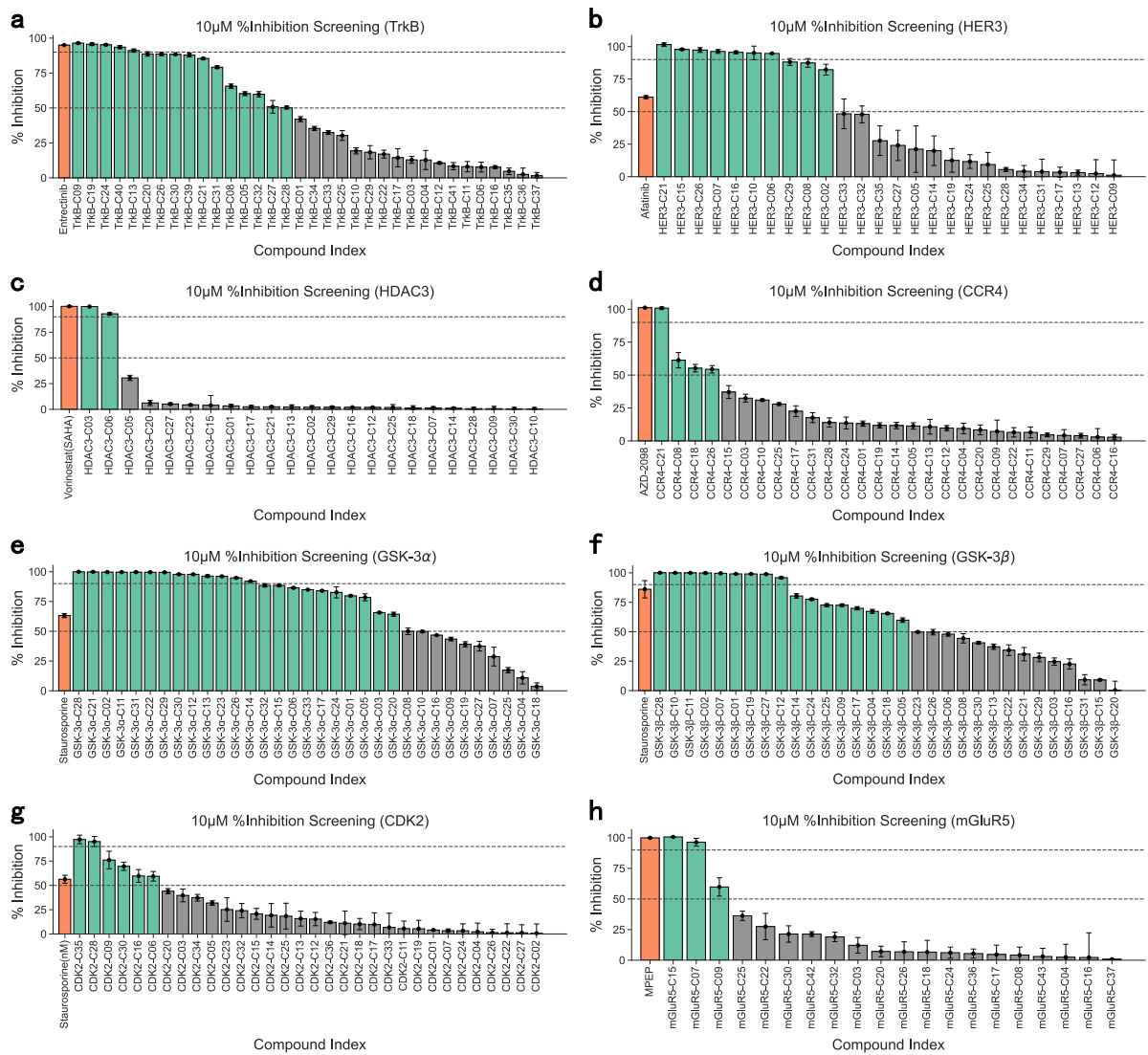

**Supplementary Fig. 3 | Inhibition distributions at 10 μM for all screened compounds across eight representative targets**. Including receptor tyrosine kinases (HER3, TrkB), G protein-coupled receptors (mGluR5, CCR4), serine/threonine kinases (CDK2, GSK-3α/β), and an epigenetic regulator (HDAC3). A substantial proportion of compounds achieved >50% inhibition at the screening concentration, supporting subsequent hit validation and characterization.

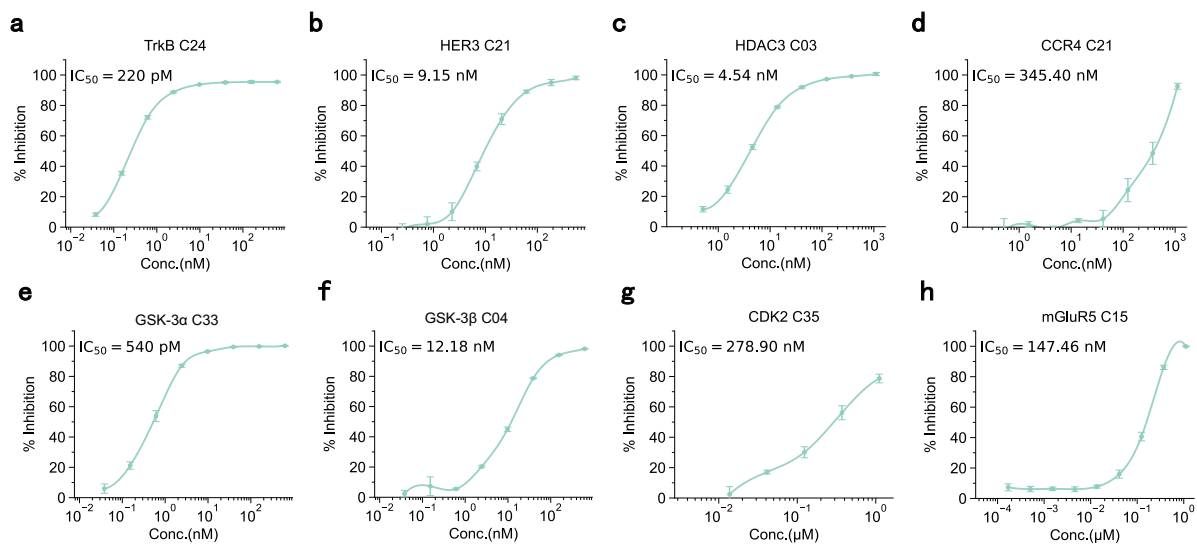

**Supplementary Fig. 4 | Dose–response validation of selected hits across eight targets.** Dose–response curves are shown for the most potent compound identified for each target based on secondary $IC_{50}$ measurements. Targets include receptor tyrosine kinases (TrkB, HER3), G protein-coupled receptors (mGluR5, CCR4), serine/threonine kinases (CDK2, GSK-3α/β), and the epigenetic regulator HDAC3. Full sets of dose–response profiles for additional validated compounds are available in the Supplementary Information.

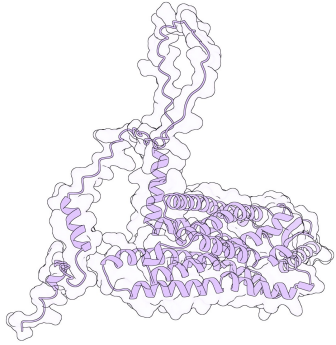 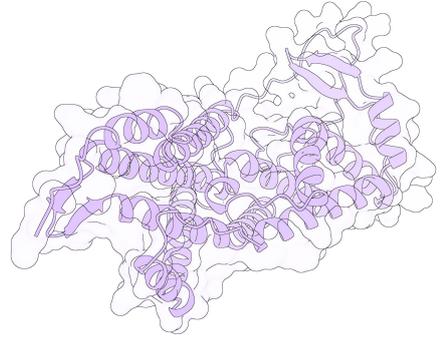

**Supplementary Fig. 5 | AlphaFold Protein Structure Database (AF-DB) models of GPR151 and GPR160.** a, GPR151 adopts a canonical class-A GPCR fold. b, GPR160 displays an atypical, conformationally flexible architecture.

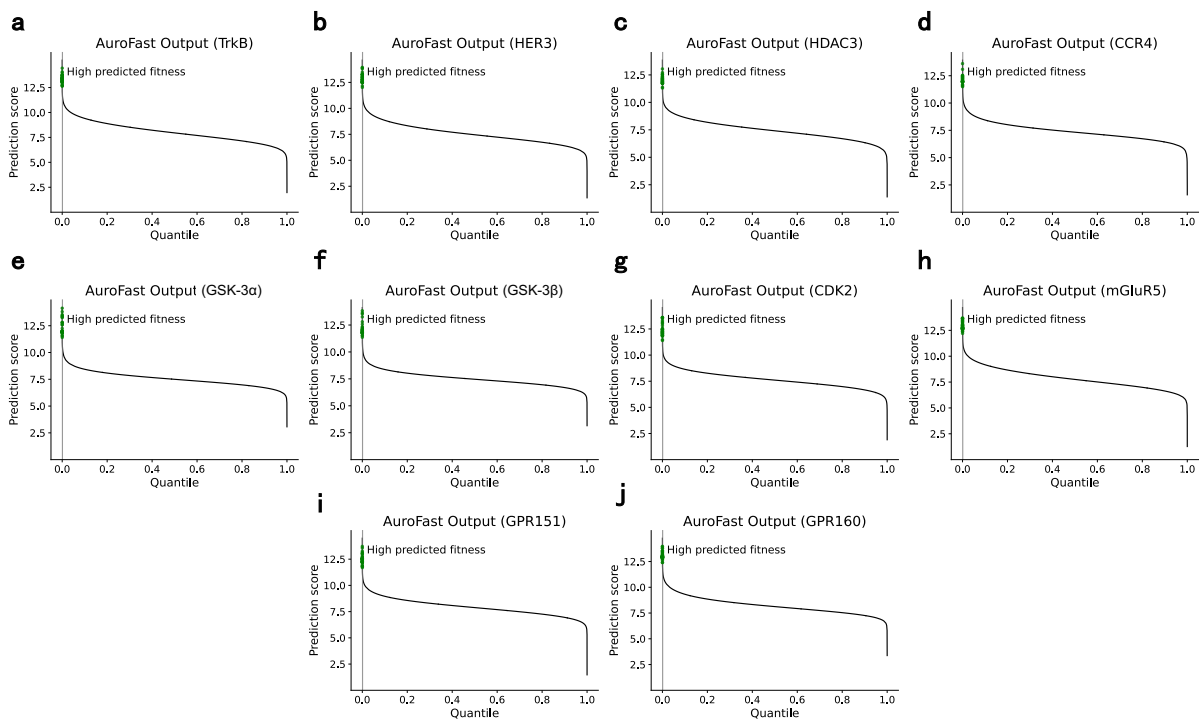

**Supplementary Fig. 6 | Predicted fitness score distributions from AuroFast virtual screening.** Quantile plots show the distribution of predicted fitness scores for compounds screened against ten different targets using the AuroBind model: (a) TrkB, (b) HER3, (c) HDAC3, (d) CCR4, (e) GSK3α, (f) GSK3β, (g) CDK2, (h) mGluR5, (i) GPR151, and (j) GPR160. Compounds with the high predicted fitness scores are shown on the left side of each plot. Green dots represent the top-ranked compounds selected for experimental validation.

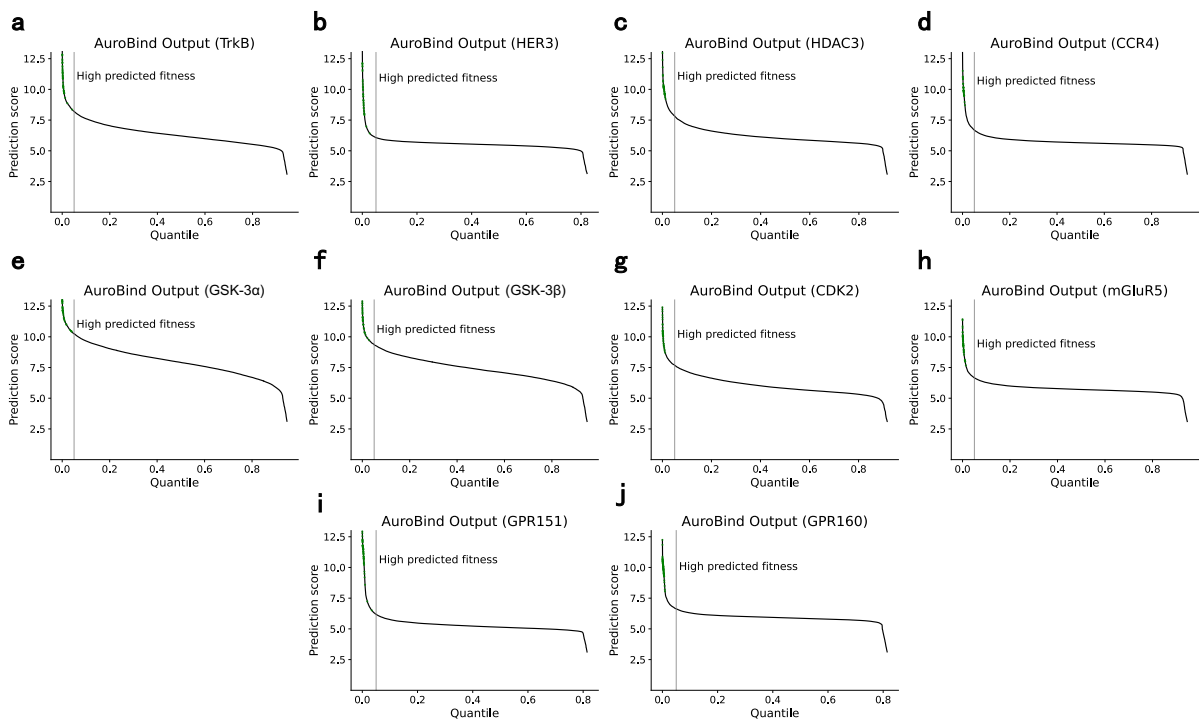

**Supplementary Fig. 7 | Predicted fitness score distributions from AuroBind virtual screening.** Quantile plots show the distribution of predicted fitness scores for compounds screened against ten different targets using the AuroBind model: (a) TrkB, (b) HER3, (c) HDAC3, (d) CCR4, (e) GSK3α, (f) GSK3β, (g) CDK2, (h) mGluR5, (i) GPR151, and (j) GPR160. Compounds with the high predicted fitness scores are shown on the left side of each plot. Green dots represent the top-ranked compounds selected for experimental validation.

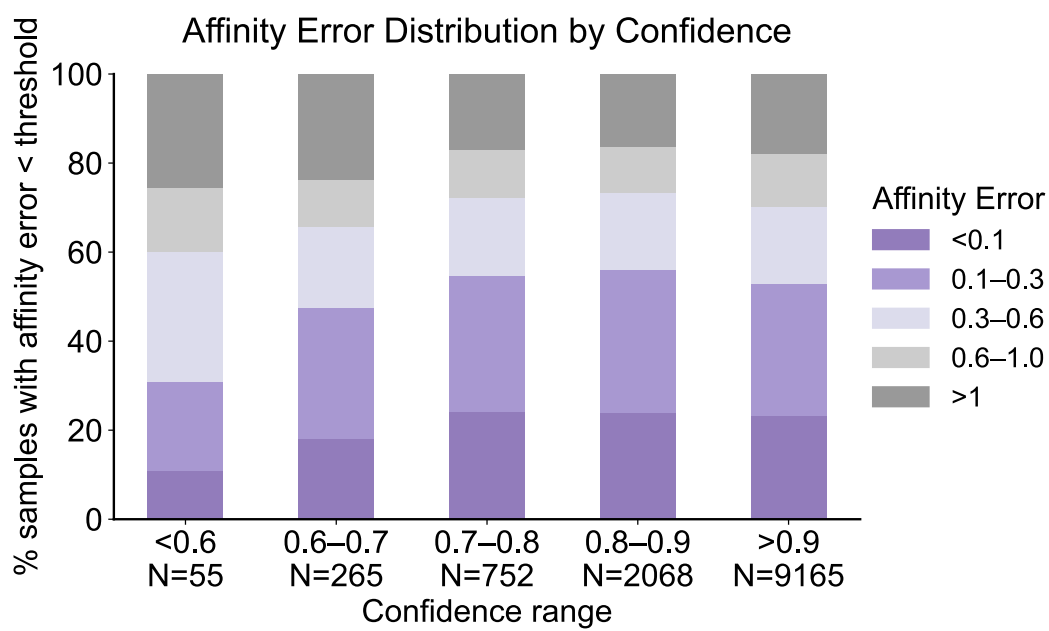

**Supplementary Fig. 8 | Influence of target protein confidence on affinity prediction accuracy.** Stacked bar charts show the cumulative percentages of compound–target pairs whose absolute affinity error falls below increasing thresholds (dark to light shading) across different predicted confidence bins (0.6, 0.7, 0.8, 0.9, and >0.9) in the test set of BindingDB benchmark. When the predicted confidence is low (<0.6), affinity prediction errors are significantly amplified, indicating that poor global structure quality substantially impairs binding affinity prediction. As the predicted confidence improves beyond 0.8, the correlation between confidence and affinity error weakens, suggesting that once an overall reliable structure is achieved, other factors such as binding pocket modeling quality or the model's perception of binding fitness become the dominant determinants of affinity prediction accuracy.

**Supplementary Table 4 | Performance of AuroBind versus the best literature-reported virtual screening Method on specific target.** For eight pharmaceutically relevant targets we compare our AuroBind (upper block) with the most successful virtual screening study reported for each target in the primary literature (middle block). For every target the first line gives the number of compounds that were experimentally confirmed as active (bold), the line in parentheses reports the total number of molecules taken forward to the assay (that is, the size of the prospective screen), and the third line lists the most potent biochemical affinity obtained in that campaign ($IC_{50}$). The bottom line indicates whether the primary hits were subjected to additional iterative wet-lab optimization cycles in the original study. AuroBind retrieves low-nanomolar to sub-nanomolar binders from prospective sets two to four orders of magnitude smaller than those used in previous work, yielding markedly higher hit rates while requiring no subsequent medicinal-chemistry optimization.

|  | CCR4 | CDK2 | GSK3-α | GSK3-β | mGluR5 | TrkB | HER3 | HDAC3 |
|---|---|---|---|---|---|---|---|---|
| AuroBind | **4** <br> (31) <br> **345nM** | **6** <br> (37) <br> **280nM** | **23** <br> (33) <br> **540pM** | **17** <br> (31) <br> **1.32nM** | **3** <br> (44) <br> **150nM** | **16** <br> (41) <br> **220pM** | **10** <br> (36) <br> **9.20nM** | **2** <br> (30) <br> **4.54nM** |
| Other screening Methods[52-58] | 15 <br> (116) <br> 2.34 μM | 10 <br> (20000) <br> 87 nM | 1 <br> (70) <br> 2.26 μM | 2 <br> (70) <br> 4.23 μM | 9 <br> (194,563) <br> 2.3 μM | 7 <br> (3,000,000) <br> 0.18 μM | 428 <br> (107, 008) <br> 4 μM | 3 <br> (212,531) <br> 1.3 μM |
| Optimized in wet-lab cycles | Yes[83] | No[84] | No[85] | No[85] | No[86] | No[87] | No[88] | Yes[89] |